\PassOptionsToPackage{dvipsnames, pdftex}{xcolor}
\documentclass[runningheads]{llncs}
\usepackage{graphicx}

\usepackage{tikz}
\usepackage{comment}
\usepackage{amsmath,amssymb} 
\usepackage{color}
\usepackage{booktabs}
\usepackage[pagebackref,breaklinks,colorlinks,citecolor=blue,linkcolor=blue,bookmarks=false]{hyperref}

\usepackage{cite}
\usepackage{microtype}
\usepackage{stackengine}

\usepackage{tablefootnote}

\usepackage[normalem]{ulem}

\usepackage[export]{adjustbox}

\usepackage[accsupp]{axessibility}  

\usepackage{setspace}
\usepackage[dvipsnames]{xcolor}

\usepackage[symbol]{footmisc}
\renewcommand{\thefootnote}{\fnsymbol{footnote}}

\usepackage{enumitem}
\usepackage{multirow}
\usepackage{wrapfig}
\usepackage{colortbl,xcolor}

\let\llncssubparagraph\subparagraph
\let\subparagraph\paragraph
\usepackage[compact]{titlesec}
\let\subparagraph\llncssubparagraph
\titlespacing{\section}{0pt}{3ex plus .2ex minus 1ex}{2ex minus 1ex}
\titlespacing{\subsection}{0pt}{2ex plus .2ex minus 1ex}{1ex minus 1ex}

\setlist[itemize]{align=parleft,left=0pt}

\definecolor{azure(colorwheel)}{rgb}{0.0, 0.5, 1.0}
\definecolor{nicegreen}{rgb}{0.0, 0.7, 0.1}
\definecolor{clova}{rgb}{0.24, 0.63, 0.33}
\definecolor{customgray}{rgb}{0.9, 0.9, 0.9}
\definecolor{pink}{cmyk}{0, 0.7808, 0.4429, 0.1412}
\definecolor{amethyst}{rgb}{0.6, 0.4, 0.8}
\definecolor{black}{rgb}{0.0, 0.0, 0.0}
\definecolor{white}{rgb}{1.0, 1.0, 1.0}
\definecolor{red}{rgb}{0.9, 0.0, 0.}

\newcommand{\rred}[1]{\textcolor{RoyalBlue}{#1}}
\newcommand{\blue}[1]{\textcolor{OrangeRed}{#1}}

\newcolumntype{g}{>{\columncolor{customgray}}c}
\newcolumntype{z}{>{\columncolor{customgray}}l}
\newcolumntype{?}[1]{!{\vrule width #1}}

\newcommand{\kjs}[1]{\textcolor{black}{#1}}
\newcommand{\ug}[1]{\textcolor{black}{#1}}
\newcommand{\moon}[1]{\textcolor{black}{#1}}

\newcommand{\ugkim}[1]{\textcolor{black}{#1}}

\renewcommand{\paragraph}[1]{\vspace{1mm}\noindent\textbf{#1.}\,\,}
\newcolumntype{C}{>{\centering\arraybackslash}p{3em}}

\def\ourpsnr{31.58}

\usepackage{xspace}

\def\onedot{.\@\xspace}
\def\eg{\emph{e.g}\onedot} 

\def\ie{\emph{i.e}\onedot}

\def\wrt{\emph{w.r.t}\onedot} 

\def\etal{\emph{et al}\onedot}

\newcommand{\Sref}[1]{Sec.~\ref{#1}}

\newcommand{\Fref}[1]{Fig.~\ref{#1}}
\newcommand{\Tref}[1]{Table~\ref{#1}}

\newcommand{\br}{{\mathbf{r}}}

\newcommand{\calT}{{\mathcal{T}}}

\newcommand{\be}{\begin{eqnarray}}
\newcommand{\ee}{\end{eqnarray}}
\newcommand{\bee}{\begin{eqnarray*}}
\newcommand{\eee}{\end{eqnarray*}}

\newcommand{\matrixb}{\left[ \begin{array}}
\newcommand{\matrixe}{\end{array} \right]}   

\begin{document}
\pagestyle{headings}
\mainmatter
\def\ECCVSubNumber{3703}  

\title{HDR-Plenoxels: Self-Calibrating\\High Dynamic Range Radiance Fields} 

\titlerunning{HDR-Plenoxels}

\author{Kim Jun-Seong\inst{1}\thanks{Authors contributed equally to this work.} \quad
Kim Yu-Ji\inst{2\ast} \quad
Moon Ye-Bin\inst{1} \quad
Tae-Hyun Oh\inst{1, 2,}\thanks{Joint affiliated with Yonsei University, Korea.}}

\authorrunning{K. Jun-Seong et al.}

\institute{
${}^1$Dept. of Elect. Eng. \qquad ${}^2$Grad. School of AI
\\Pohang University of Science and Technology (POSTECH)\\
\email{\{junseong.kim, ugkim, ybmoon, taehyun\}@postech.ac.kr}\\
\url{https://github.com/postech-ami/HDR-Plenoxels}
}

\maketitle

\begin{abstract}

We propose high dynamic range (HDR) radiance fields,
HDR-Plenoxels, that learn a plenoptic function of 3D HDR radiance fields, geometry information, and varying camera settings inherent in 2D low dynamic range (LDR) images.
Our voxel-based volume rendering pipeline reconstructs HDR radiance fields with only multi-view LDR images taken from 
varying camera settings in an end-to-end manner and has a fast convergence speed.
To deal with various cameras in real-world scenario\ugkim{s}, we introduce a tone mapping module that models 
the digital in-camera imaging pipeline (ISP) and disentangles radiometric settings.
Our tone mapping module allows us to render by controlling the radiometric settings of each novel view.
Finally, we build 
a multi-view dataset with varying camera conditions, which fits our problem setting.
Our experiments show that HDR-Plenoxels can express detail 
and high-quality 
HDR novel views from only LDR images with various cameras.

\end{abstract}
\section{Introduction}
The human eyes can respond to a wide range of brightness in the real-world scene, from very bright to very dark, \ie, high dynamic range (HDR).
\ugkim{The human} can see an object with its color and texture even in dark and dim conditions.
However, standard
digital cameras capture 
a limited range of scenes 
due to 
the low dynamic range (LDR) limits of the sensors.
HDR imaging and display techniques
 have been developed to overcome the sensors' limits and to share the beauty of the world as humans see.

Existing studies on HDR recovery~\cite{debevec_1996_modeling,exp_fusion,Robertson99dynamicrange,sen_2012_robust} have been mainly focused on a static view HDR from a monocular perspective or HDR video recovery.
HDR images are typically
reconstructed by merging multi-exposure LDR images in a fixed camera pose.
To recover an HDR image from LDR images taken from various viewpoints, the prior work~\cite{sen_2012_robust,chen_hdr_2021} 
suggests accumulating images after alignment.
However, the HDR reconstruction results of the prior work are still limited to a given view. 
To overcome the limitations, we propose a method of restoring HDR radiance fields with only multi-view LDR images.
The LDR images taken from varying cameras are used, 
where 
various radiometric conditions exist, including 
different exposure, white balance, and camera response functions (CRFs).

\begin{figure}[t]
    \centering
    \includegraphics[width=1\linewidth]{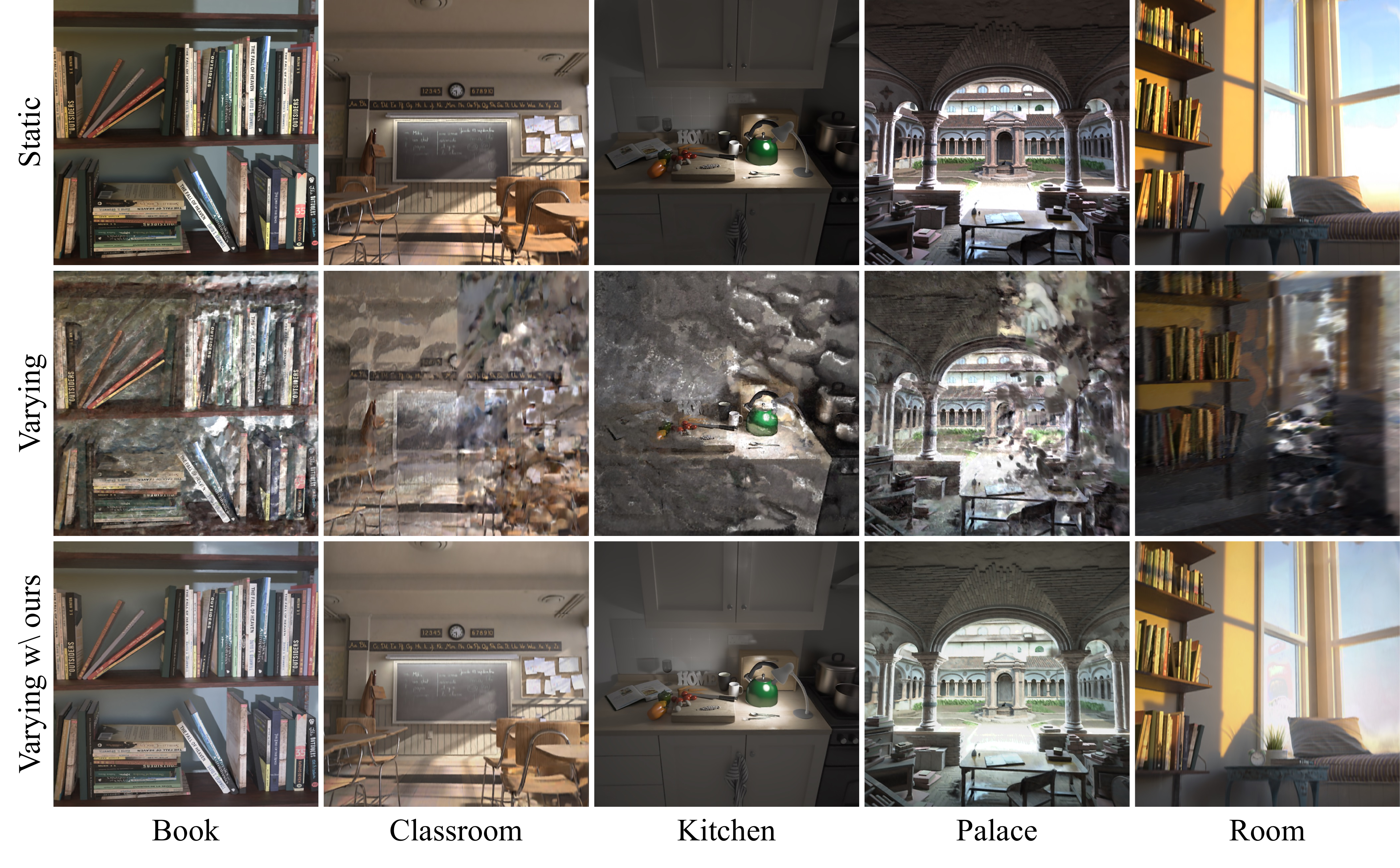}
    \caption{\textbf{Qualitative results of static and varying camera settings.}
    The static and varying mean camera conditions include exposure, white balance, and CRF.
    The static camera condition is a controlled environmental setting, \ie, all views of the scene have the same components of exposure, white balance, and CRF.
    The varying camera condition is alternated environmental settings, \ie, all views of the scene have different components.
    Each row represents camera conditions, and each column represents the class of synthetic datasets.}
    \label{fig:qual_in_ours}
\end{figure}

The novel view synthesis requires additional information to reconstruct and synthesize unseen views,
given a sparse set of images.
Previous arts inject 
prior knowledge by
voxel-based~\cite{liu_neural_2021, yu_plenoxels_2021, yu_plenoctrees_2021}, mesh-based~\cite{waechter_let_2014, buehler_unstructured_2001, wood_surface_2000, debevec_1996_modeling}, multi-plane~\cite{Wizadwongsa2021NeX}, and volume rendering~\cite{nerf, lombardi_neural_2019} 
to cope with the problem.
Recently, 
Plenoxels~\cite{yu_plenoxels_2021} have shown outstanding efficiency of the voxel-based method by assigning 
spherical harmonics to each voxel corner. 
While maintaining comparable qualitative results, the Plenoxels achieve two orders of magnitude faster rendering speed than the Neural Radiance Fields (NeRF)~\cite{nerf}, which utilizes the implicit neural functions to conduct 
volume rendering.
In this work, we extend Plenoxels by proposing HDR-Plenoxels that can restore HDR radiance fields with LDR images under diverse camera conditions in an end-to-end manner.
Although many saturated regions are appeared due to a wide dynamic range of a scene 
during training, our HDR-Plenoxels are robust to saturated regions and represent accurate geometry and color at rendering.
This is achieved by proposing 
a tone mapping module that approximates the in-camera pipeline, camera \kjs{image signal processings (ISPs)}, 
from HDR radiance to LDR intensity, allowing flexible modeling 
of various radiometric and non-linear camera conditions.
The tone mapping parameters are spontaneously learned during training.
In addition, once 3D HDR radiance fields and the CRFs are fitted, our tone-mapping module can be freely controlled to synthesize different radiometric conditions of rendering in any 
view.
Our tone mapping module can easily be attached to most volume rendering model variants as well.


Our HDR-Plenoxels mainly consist of two parts:
1) HDR radiance fields modeled by Plenoxels followed by 2) the tone mapping module.
The differentiable tone mapping module renders HDR radiance values composited from Plenoxels into LDR intensity, which allows to back-propagate gradients to the voxel grid, so that spherical harmonics (SH) coefficients and opacity are learned to span the HDR radiance with the scene geometry.
The tone mapping module explicitly models CRFs, which enables to self-calibrate CRFs of each view during training. In addition, thereby, we can easily edit the rendering property by just controlling the radiometric curves of each novel view by virtue of the disentangled parameterization of the  module.
Our experiments show that our method achieves preferable performance on novel view synthesis with varying radiometric conditions of input.
Our main contributions are summarized as follows:
\begin{itemize}
    \item We propose
    an end-to-end HDR radiance field learning method, HDR-Plenoxels, that allow learning 3D HDR radiance fields from only
    multi-view and varying radiometric conditioned LDR images as input.
    \item We model the tone mapping module based on a physical camera imaging pipeline that maps HDR to LDR with 
    explicit radiometric functions.
    \item We build 
    a multi-view dataset containing varying camera conditions. 
    The dataset includes synthetic and real scenes with various camera settings such as white balance, exposure, and CRF.
\end{itemize}



\section{Related Work}
The scope of our work contains HDR imaging, volume rendering, and its calibration.
We overview the prior work \ugkim{from} each perspective in this section.

\paragraph{HDR Imaging}
A standard 
HDR recovery~\cite{debevec_1996_modeling} directly accumulates
multi-exposure LDR images taken from a fixed camera pose, which are prone to ghosting artifacts in dynamic scenes.
To cope with the limitation, several studies~\cite{sen_2012_robust,chen_hdr_2021} suggest
a method to recover an HDR image from LDR images taken from moving cameras by using image alignment methods, such as image warping or optical flows.
However, they still suffer from large camera motion and occlusion due to the imperfect warping model in the alignment step. 
In contrast, our work exploits multi-view geometric information, which enables to obtain radiometrically calibrated HDR radiance of
an entire 3D scene and to be
robust even with large camera motion and occlusion.

Typical digital cameras can only deal with LDR due to the limited dynamic range and the inherent nonlinear components, which represent the real-world scene irradiance inferiorly in pixel values and cause discrepancy to the real scene during the image processing~\cite{debevec_1996_modeling}.
To obtain an accurate HDR, we have to understand an inherent nonlinear relationship of the camera, \ie, the radiometric properties of camera \kjs{ISPs}.
Traditional radiometric calibration models the components of physical pipelines of cameras, including white balance and CRF, and optimizes to reconstruct HDR from only given LDR images~\cite{debevec_1996_modeling} or HDR-LDR image pairs~\cite{kim_new_2012}.
The latest learning-based approaches~\cite{eilertsen_2017_hdr,endo_deep_2017,marnerides_expandnet_2019} suggest
an implicit model-based method but 
require ground truth HDR images paired with LDR images for training.
Liu~\etal~\cite{liu_single-image_2020} replace an implicit function with an explicit physical camera model, enhancing the HDR image reconstruction quality.
Our method shares the same advantages by adopting the explicit tone-mapping module.
Note that our HDR-Plenoxels learn HDR radiances up to scale, given only LDR images but without ground truth HDR images and camera parameters.

\paragraph{Volume Rendering}
Volume rendering is the method of understanding the 3D information inherited in two-dimensional images to render images at unseen views, called novel views.
\moon{Existing methods~\cite{nerf,sitzmann_scene_2020} show high performance in complex geometric shapes but require high memory for high expressiveness.}

The recent volume rendering methods utilize multi-layer perceptron (MLP) based implicit neural function to predict the signed distance fields~\cite{park_deepsdf_2019,gropp_implicit_2020,yariv_multiview_2020} and occupancy~\cite{mescheder_occupancy_2019,peng_convolutional_2020,saito_pifu_2019}, and demonstrate the high expressiveness with high compression power.
In particular, Neural Radiance Field (NeRF)~\cite{nerf} shows fine-detailed rendering performance unprecedently.
However, the NeRF-related studies~\cite{wang_nerf--_2021,martin-brualla_nerf_w_2021,park_nerfies_2021,pumarola_d-nerf_2020} have a limitation of high training and rendering time complexity due to the forward process in every sampling point.
Several studies~\cite{reiser_kilonerf_2021,neff_donerf_2021,song_autoint_2019,hedman_baking_2021} try to modify the neural network to reduce the computation at each sampling point to reduce the time cost.

Octree structure-based methods~\cite{liu_neural_2021,yu_plenoctrees_2021,yu_plenoxels_2021} are efficient methods that reduce rendering time by virtue of 
their structure.
Plenoxels~\cite{yu_plenoxels_2021} optimize the octree structure with spherical harmonics instantaneously, requiring
only 
tens of 
minutes of training time 
to achieve detailed rendering results comparable to NeRF.
Our method uses Plenoxels as a volume rendering backbone for efficient rendering, and further expands the expression power of Plenoxels to 3D HDR radiance fields with negligible computational cost.

\paragraph{Calibrated Volume Rendering}
Several methods~\cite{mildenhall_nerf_2021,huang_hdr-nerf_2021,martin-brualla_nerf_w_2021,ruckert_adop_2021} are proposed for better performance with relaxed assumptions in volume rendering.
NeRF in the Wild (NeRF-W)~\cite{martin-brualla_nerf_w_2021} uses web images in the wild setting for reconstruction, which deals with varying camera conditions and occluded objects by introducing a handling mechanism by appearance and transient embeddings.

ADOP~\cite{ruckert_adop_2021} is a point-based HDR neural rendering pipeline that consists of a differentiable and physically-based tone mapping. 
Due to the differentiable property, all the varying camera conditions can be optimized.
However, ADOP requires dense COLMAP structure-from-motion package~\cite{schonberger_structure-motion_2016} results as input, which have expensive time costs.
Also, the method itself has expensive
time and memory costs during training due to the point-based method.
Distinctive from ADOP, our method is cost efficient by utilizing an octree-based structure and camera poses without dense COLMAP.

Recently, HDR-NeRF~\cite{huang_hdr-nerf_2021} tackles a similar HDR radiance field problem with ours by NeRF, which is concurrent work with us. 
The work requires known exposure information and does not take into account white balance parameters, in contrast to ours.








\section{HDR Radiance Fields from Multi-view LDR Images}

\begin{figure}[t]
    \setlength{\abovecaptionskip}{1mm} 
    \setlength{\belowcaptionskip}{-2mm} 
    \centering
    \includegraphics[width=1\linewidth]{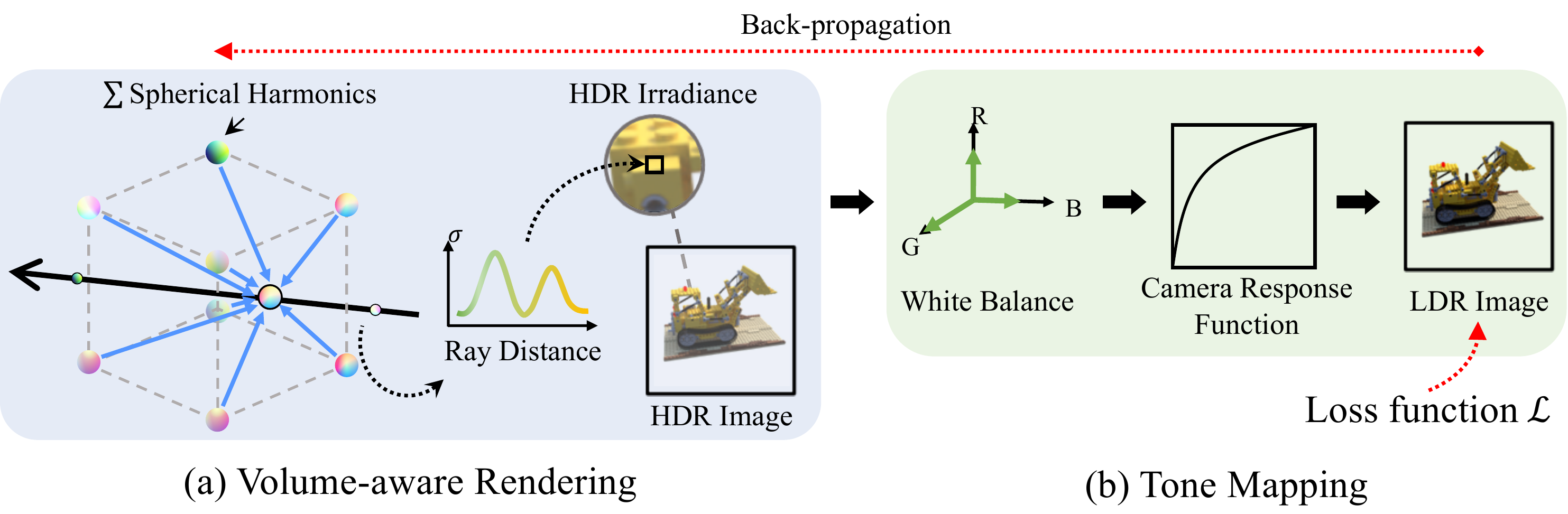}
    \caption{\textbf{Overall pipeline of HDR-Plenoxels.} 
    1) Plenoxels synthesize an HDR image from HDR radiance by ray-marching, then 2) the differentiable tone-mapping function maps from HDR to LDR in an end-to-end manner. 
    The self-calibration is done by minimizing 
    the residual between 
    the synthesized LDR image and the captured one
    with regularizations.
    }
    \label{fig:pipeline}
\end{figure}

Our work aims to reconstruct HDR radiance fields of a visual scene from multi-view LDR input images. 
In this section, we first present the overall pipeline of our method~(\Fref{fig:pipeline}), which is composed of two parts, 1) volume-aware HDR image rendering (\Sref{sec:3.1}), and 2) synthesis of LDR images through the tone mapping module (\Sref{sec:3.2}). 
We then explain the details of optimization (\Sref{sec:3.3}).


\subsection{Volume-aware Rendering to HDR Images}\label{sec:3.1}
To reconstruct the HDR irradiance fields from the multi-view LDR images, we parameterize HDR radiance fields 
by voxel grids with spherical harmonics (SH)
called 
Plenoxels~\cite{yu_plenoxels_2021}.
A bounded three-dimensional space of interest 
is represented as a sparse voxel grid, each of which has opacity and SH coefficients.
The volume rendering method adopts a coarse-to-fine training scheme similar to NSVF~\cite{liu_neural_2021}.
The learning process starts with a broad and uniformly divided sparse voxel grid, and the voxels are upsampled to make a denser grid as learning progresses.
The voxels are pruned according to the occupancy threshold to reduce the computational cost, as training iterations go.
Upsampling and pruning are applied simultaneously during training and repeated several times.
%


As SH 
can compactly represent  
any functions on a sphere well
with 
\ug{a few SH bases~\cite{basri_lambertian_2003, yu_plenoctrees_2021},}
it has been vastly used in graphics for HDR environmental lighting~\cite{ravi_sh_env_01} and glossy representation~\cite{Mahajan:2008:ATF}, which motivates our HDR radiance representation. 
Furthermore, recent work has adopted SH in volume contents \ug{and} has demonstrated its effectiveness in implicitly expressing the non-Lambertian effects~\cite{Wizadwongsa2021NeX,kuang2021neroic}.
Our HDR irradiance field modeling by SH based volume-aware rendering 
exploits these advantages.

The SH in each voxel grid is used for view representation.
All the colors over every direction of a sphere 
are spanned by 
a weighted sum of pre-defined spherical basis functions and coefficients corresponding to each function.
Therefore, the corresponding color can be defined for each specific angle.
Each vertex of the voxel grid stores 28-dimensional vectors: 27 for SH coefficients (9 coefficients per color channels) and 1 for voxel opacity $\sigma$.
Empirically, the values of the SH coefficients change significantly during training which makes the training unstable.
To mitigate this, we initialize the color to grey by adding \ugkim{the} offset color 0.5
and 
the voxel opacity value to 0.1.

The HDR volume rendering part determines the color of a \ug{rendered HDR} image pixel $\hat{C}(\mathbf{r})$ by ray-marching the color and opacity of points sampled along a ray in a bounded three-dimensional voxel grid volume.
At any 3D point $\mathbf{(x,y,z)}$ and normalized viewing angle $\mathbf{(v_{x}, v_{y}, v_{z})}$ inside the voxel grid of Plenoxels, the color and opacity of the point are trilinearly interpolated from eight nearby voxel vertices.
For a camera center $\mathbf{o}$ and given an image pixel grid, we can define the ray $\mathbf{r}=\mathbf{o}+t\mathbf{d}$ starting from $\mathbf{o}$ to each pixel in the camera along the direction $\mathbf{d}$.
After that, $N$ sampling points are sampled over the ray at regular intervals $\delta_{i}=t_{i+1}-t_{i}$. 
The color and opacity of each sampled point are denoted as  $\mathbf{c_{i}}$ and $\sigma_{i}$, respectively, and $T_{i}$ \ugkim{is}
the accumulated transmittance value up to the $i$-th point.
The ray-marching proceeds as follows:
\begin{equation} \label{eq:1}
    \hat{C}(\mathbf{r})=\sum_{i=1}^{N} T_{i}\left(1-\exp \left(-\sigma_{i} \delta_{i}\right)\right) \mathbf{c}_{i},       \ \textrm{where} \ T_{i}=\exp \left(-\sum_{j=1}^{i-1} \sigma_{j} \delta_{j}\right).
\end{equation}
The ray sampling randomly selects 
rays among the set of rays toward all pixels of an image for efficient training.


\subsection{Tone Mapping}\label{sec:3.2}
\ug{The tone mapping stage converts an HDR image into an LDR image.
We denote a pixel value of an HDR image and an LDR image as $I_{h}$ and $I_{l}$, respectively.}
The output of volume rendering is HDR radiance fields, and we obtain $I_{h}$ as output by ray marching HDR radiance fields. 
We represent our explicit tone mapping module 
as a function $\mathcal{T}$ with radiometric parameters $\theta$, \ie, 
$I_{l} = \mathcal{T}(I_{h}, \theta)$.

The tone mapping function $\mathcal{T}$ consists of two stages.
Each stage is parameterized by the physical property of its components and represented as separate functions: white balance function $w$ and camera response function (CRF) $g$.
Note that we regard the white balance scale parameters are merged with the exposure value and learned at once.
Two sub functions are applied sequentially as $I_{l} = \mathcal{T}(I_{h}) = g\circ (w (I_{h}))$, which follows 
the image acquisition process of common digital cameras.

Specifically, first, for a specific ray $\textbf{r}$, the pixel color of an HDR image $C_{h}(\textbf{r})$ is calculated through ray-marching.
The white balance function $w(\cdot)$ is applied to $C_{h}(\textbf{r})$ with $\theta_{w} = [w_{r}, w_{g}, w_{b}]^{\top} \in \mathbb{R}^3$, and the function output is \ug{a pixel of} white balance calibrated image $I_w$. 
That is, given each channel components of $C_h(\mathbf{r})=[c_{h}^{r}, c_{h}^{g}$,$c_{h}^{b}] \in \mathbb{R}^3$, 
\begin{equation}
    I_{w}=
    w(C_{h}(\textbf{r}), \theta_{w})
    =
    {C_{h}(\textbf{r})} 
    \odot 
    \theta_{w}
    = 
    \left[\begin{array}{c}
    c_{h}^{r} \\
    c_{h}^{g} \\
    c_{h}^{b}
    \end{array}\right]\left.
    \odot 
    \left[\begin{array}{l}
    w_{r} \\
    w_{g} \\
    w_{b}
    \end{array}\right]\right.
    =
    \left[\begin{array}{c}
    w_{r} c_h^{r} \\
    w_{g} c_h^{g} \\
    w_{b} c_h^{b}
    \end{array}\right],
\end{equation}
where the operator $\odot$ stands for an element-wise product.
To make white balance physically proper, we regularize $\theta_{w}$ to be a positive value.




The CRF $g$ is applied to $I_{w}$.
We parameterize non-linear CRFs with an approximated discrete piece-wise linear function.
The function $g(\cdot)$ is divided into 
256 intervals 
which are allocated for uniformly sampled points in [0,1], and parameterized by 256 control points. 
The \ug{pixel value of} white balance corrected image $I_{w}$ is mapped to $I_{l}$ by interpolating corresponding CRF values of nearby control points in \ugkim{the} domain.
To make the CRF $g(\cdot)$ differentiable, we adapt \ugkim{the} 1D grid-sampling used in~\cite{jaderberg_spatial_2016}.
According to Debevec~\etal~\cite{debevec_1996_modeling}, the CRF is enforced to follow 
the following boundary condition: $I_{l}=g\left(I_{w}; \theta_{g}\right)$, $g(0) = 0$, $g(1)=1$.
A range beyond the dynamic range [0,1] is thresholded when applying the CRF $g$.
To propagate a loss \ugkim{in} the saturation region of the rendered images during training HDR radiance fields, we apply the leaky-thresholding method:
\begin{equation}
    g_{leaky}(x)=\left\{\begin{array}{cc}
    \alpha x ,  & x<0 \\ 
    \operatorname{g}(x) ,  & 0 \leq x \leq 1 \\
    -\frac{\alpha}{\sqrt{x}}+\alpha+1 , & 1<x,
    \end{array}\right.
\end{equation}
where $\alpha$ is the thresholding coefficient.

\subsection{Optimization}\label{sec:3.3}

\noindent\begin{wrapfigure}{r}{0.42\linewidth}
  \centering
  \includegraphics[width=0.8\linewidth]{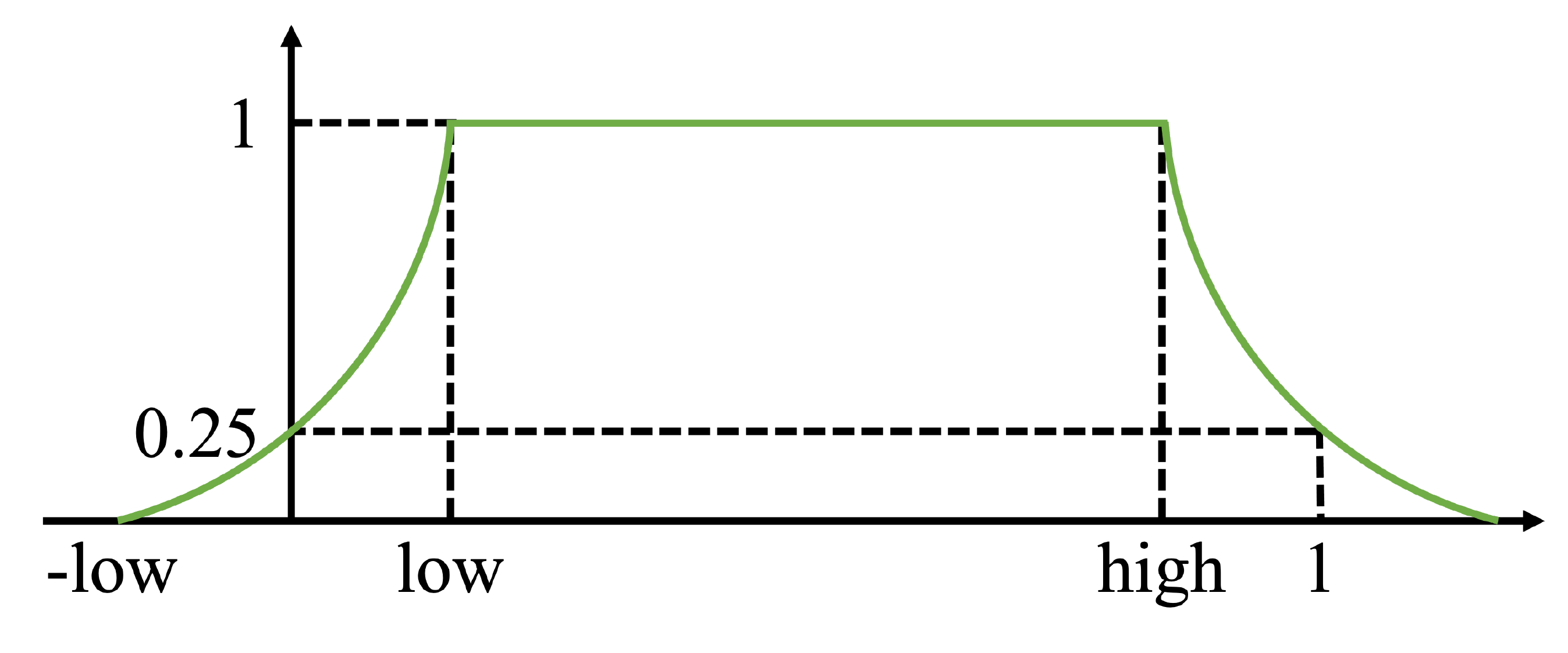}
  \caption{\textbf{Leaky saturation mask.}}
  \label{fig:saturation mask}
\end{wrapfigure}\textbf{Leaky Saturation Mask.}\quad\label{sat}
When taking a scene with a wide dynamic range, the LDR images may contain over- and under-saturation.
In the saturated region, no cue exists to guess
correct geometric and photometric information due to missing texture.
This acts as outliers when taking into account the loss computation during optimization.
To suppress the impact of saturation regions and 
\ug{prevent our recovery from being biased,}
we use saturation masking in the loss computation (\Fref{fig:saturation mask}).
We define our leaky saturation mask 
as follows:
\begin{equation}
    \operatorname{\mathtt{mask}}(x)
    =
    \left\{
    \begin{array}{cc}
    \left(\tfrac{x+\mathsf{low}}{2\mathsf{low}}\right)^{2} & x < \mathsf {low}, \\[1mm]
    1, & \mathsf{low} \leq x \leq\mathsf {high}, \\
    \left(\tfrac{2-x}{2(1-\mathsf{high})}\right)^{2} & x > \mathsf{high},
    \end{array}\right.
\end{equation}
We empirically set $\mathsf{low}=0.15$ and $\mathsf{high}=0.9$ for the experiment.


\paragraph{White Balance Initialization}\label{init}
There exists inherent ambiguity in the camera imaging pipeline, which occurs due to the inherent entangled relationship across 
model components in the pipeline, \eg, for an image of a view, if we increase its exposure time twice while reducing the white balance of the view by half, the resulting image appears same with the original setting of exposure time; \ie, there are multiple solutions that can produce the same LDR images.

We avoid such cumbersome 
ambiguity between exposure and white balance in our method.
We use only the white balance module $w$ to express both exposure \ug{as} a scale and white balance ratio following the study~\cite{kim_new_2012} to workaround the scale ambiguity between exposure and white balance.
However, with this representation, 
the overall scale of the white balance is trained extremely small or large.
Therefore, we calculate the averaged color of all inputs $(r_{a}, g_{a}, b_{a}) \in \mathbb{R}^3$, select a reference image which has the closest value of $(r_{a}, g_{a}, b_{a})$, and fix the white balance of the reference image $(r_{ref}, g_{ref}, b_{ref}) \in \mathbb{R}^3$.
This acts as regularization.
This helps white balance be learned on the proper scale, which also means we have a suitable exposure value.

However, there still exists a similar ambiguity between \emph{SH coefficients} and \emph{white balance}.
We observe that when the exposure differences are significantly dynamic among neighborhood views, observed LDR intensity differences by exposure times
are misunderstood as the cause of the high-frequency reflectance\footnote{The high-frequency reflectance refers to the case that a subtle view direction change results in drastic reflectance ratio changes, such as glossy materials.} of the scene and different white balances.
This tends to produce wrong geometry as the rays have reached different parts of the scene.
We also found that the more abrupt the intensity changes among neighborhood views are, the more dominant the coefficients corresponding to high-frequency SH components become.

As a simple workaround, we use white balance information for each camera as a prior to resolve the ambiguity on the SH side.
We introduce white balance initialization for each camera that guides initial solutions to physically plausible solutions; thereby, the optimization process becomes more stable and faster to converge to desirable solutions, and robust to such harsh input conditions.
We estimate a reference color ratio by comparing per-image averaged pixel values to averaging each $rgb$ value from the entire image set $\mathcal{S}$.
Then the initial white balances $wb_{c,i}$ for each camera, where $c\in\{r,g,b\}$ of each image $I_{i}$ are initialized as 
$wb_{c,i}=\tfrac{\operatorname{mean}_{k \in I_{i}}\left(c_{k}\right)}{\operatorname{mean}_{j\in\mathcal{S}}\left(c_{j}\right)}$.


\paragraph{Spherical Harmonics Regularization}
In the harsh input condition case, where LDR images obtained from neighborhood views have significantly dynamic exposure differences, the above initialization stabilizes early optimization steps.
If the optimization speed of the white balance does not match that of the SH coefficients, 
the ambiguity may arise again in later optimization steps.
In order to regularize this, we introduce SH coefficient masking that allows scheduling to learn from diffuse reflectance property (view direction invariant radiance) first to view direction sensitive ones, \ie, low frequency order SH to high frequency ones.
We apply SH masking to the coefficient of SH of degrees 2 and 3.
We decrease the rate of SH masking by 1/5 per epoch during the early five epochs for gradual learning.
After the early five epochs, we update SH of all degrees with full rate, \ie, no SH masking.
This scheduling notably stabilizes the optimization for the harsh condition input case.

\paragraph{Loss Functions}
We optimize our pipeline \wrt voxel opacity, SH coefficients, white balance, and CRF, given multi-view LDR images as input, with the following objective function:
\begin{equation}
    \mathcal{L}=\mathcal{L}_\textrm{recon}+\lambda_\textrm{TV} \mathcal{L}_\textrm{TV}+\lambda_\textrm{smooth}\mathcal{L}_\textrm{smooth},
\end{equation}
where each term is defined as follows.
The LDR reconstruction loss for $i$-th image is defined as:
\begin{equation}
\mathcal{L}_{\text {recon }}=\frac{1}{|\mathcal{R}|} \sum\nolimits_{\mathbf{r} \in \mathcal{R}}
M_i(\mathbf{r})
\|I_i\left(\mathrm{\Pi}_i(\br)\right)
-\calT\left(\hat{C}(\mathbf{r})\right)\|_{2}^{2},
\end{equation}
where $\mathrm{\Pi}_i(\cdot)$ denotes the camera projection operator from a ray to the 2D pixel coordinate of the $i$-th LDR image, and $M_i(\mathbf{r}) = \mathtt{mask}(I_i\left(\mathrm{\Pi}_i(\br)\right))$ denotes the saturation mask computed from 
\ug{input}
LDR images.
We randomly sample rays $\br_{sampled}$
among the possible set of rays $\mathcal{R}$ from $\mathcal{N}$ images. 
The loss is calculated through a color difference between the rendered results and the ground truth LDR values along each ray $\br_{sampled}$
considering the saturation masking $M_i(\mathbf{r})$.
The total loss is applied by normalizing the number of rays sampled.
The other two terms are for regularization.
The total variation loss is defined as:
\begin{equation}
    \mathcal{L}_{T V}=\frac{1}{|\mathcal{V}|} \sum\nolimits_{\mathbf{v} \in \mathcal{V}, d \in[D]} \sqrt{\Delta_{x}^{2}(\mathbf{v}, d)+\Delta_{y}^{2}(\mathbf{v}, d)+\Delta_{z}^{2}(\mathbf{v}, d) +\epsilon},
\end{equation}
where the differences $\Delta_\cdot$ are calculated between successive voxels along each respective $(x,y,z)$-axis, \eg, the $d$-th voxel value at $(x, y, z)$ and the $d$-th voxel value at $(x+1, y, z)$ for $x$-axis.
The total variation loss is applied for opacity $\sigma$ and SH coefficients separately.
This encourages spatial and color consistency in the voxel space.
In the implementation, we use different weighting for SH coefficients $\lambda_\textrm{TV,SH}$ and opacity $\lambda_\textrm{TV,$\sigma$}$.

The smoothness loss is for obtaining a physically appropriate CRF~\cite{debevec_1996_modeling} such that CRFs increase smoothly, which is defined as: 
\begin{equation}
    \mathcal{L}_{smooth}=\sum\nolimits_{i=1}^N \sum\nolimits_{e \in [0,1]} g_i^{\prime \prime}(e)^{2},
\end{equation}
where $g^{\prime \prime}(e)$ denotes the second order derivative of CRFs \wrt the domain of CRFs.
We set $\lambda_\textrm{TV,$\sigma$}=5\cdot 10^{-4}$, $\lambda_\textrm{TV,SH}=1\cdot10^{-2}$ and $\lambda_\textrm{smooth} =1\cdot10^{-3}$.

\section{Experiments}

We compare the qualitative and quantitative results
of our HDR-Plenoxels from three perspectives: LDR image rendering accuracy, HDR irradiance image, and 3D structure reconstruction quality.
We also conduct an ablation study on our tone mapping components to demonstrate that our tone mapping components efficiently understand camera settings.

\subsection{Experimental Settings}
\kjs{Due to the lack of open datasets that fit our experiment setting, \ie, \kjs{multi-view} images under varying camera conditions, we collect synthetic and real images with various exposure, white balance, and vignetting.}
The details of the synthetic and real datasets are explained in the supplementary material.
\paragraph{Baseline and Counterparts}
We compare our proposed method against \moon{the following}
three models: original Plenoxels \kjs{(baseline)}~\cite{yu_plenoxels_2021}, NeRF-A~\cite{martin-brualla_nerf_w_2021}, and Approximate Differentiable One-Pixel Point Rendering (ADOP)~\cite{ruckert_adop_2021}.
We conduct experiments for two environmental settings: the varying camera, which has various environments that are exposure, white balance, and CRF, and the static camera, which has controlled environments.
\kjs{The baseline with inputs of the static camera is assumed to be an upper bound of the task performance.} 

\paragraph{Evaluation}
HDR-Plenoxels can learn the 3D HDR radiance fields and freely control the appearance of a novel view, \ie, unseen viewpoint, according to camera conditions.
We compute the similarity between the LDR image synthesized in the novel view and the ground truth LDR image to measure the performance for the novel view synthesis tasks.
For quantitative evaluation, we use three metrics, PSNR, SSIM~\cite{wang_multiscale_2003}, and LPIPS~\cite{zhang_unreasonable_2018}.
Our method and NeRF-A need to learn tone mapping function and appearance embedding parameters separately at each image.
We cannot predict the tone mapping parameters at novel views;
thus, we use the left half of the test image (\ie, unseen view) at training and evaluate the performance of the right half with learned parameters.
\begin{figure}[t]
\setlength{\belowcaptionskip}{-1mm} 
    \centering
    \includegraphics[width=1\linewidth]{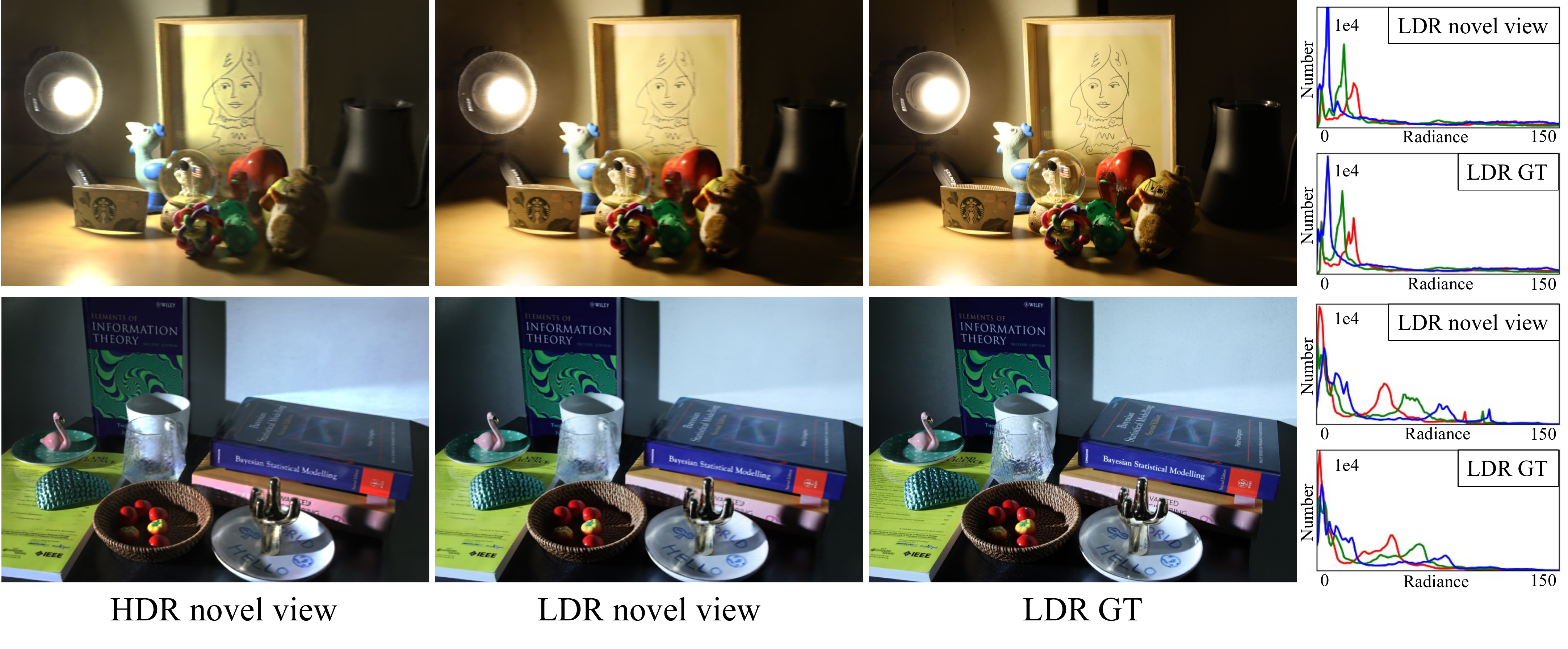}
    \caption{\textbf{Qualitative novel view results in real scenes with various camera conditions.
    }
    Each row represents different real scenes, and the column represents HDR novel view rendering, LDR novel view rendering, LDR GT (\ie, inputs of training), and RGB histograms of LDR novel view and LDR GT in order.
    Histograms of RGB are clipped from 0 to 150 at RGB radiance and from 0 to $10^{4}$ at the number of RGB radiance for visual better visibility.
    }
    \label{fig:qual_histogram}
\end{figure}

\subsection{High Dynamic Range Radiance Fields}
\moon{We evaluate the effectiveness of our HDR-Plenoxels by comparison against the counterpart models, which handle images with varying appearances.}
\begin{table}[t]
    \setlength{\belowcaptionskip}{0mm}
    \centering
    \caption{\textbf{Quantitative results of novel view synthesis on synthetic data.}
    Values are \ugkim{the} average of the results for the test data of each scene.
    For evaluation, ADOP exploits all the input images as it needs dense reconstruction.
    For the other models, the left half of the image was included in the learning data and learned, and tested on the unseen right half of the image.
    Our model shows overall high performance compared to other models.
    $\mathcal{S}$ denotes the static, and $\mathcal{V}$ is the varying datasets.
    The blue and red colors stand for the \textbf{\rred{best}} and the \textbf{\blue{second best}}, respectively.
    }
    \resizebox{\textwidth}{!}{ 
        \begin{tabular}{c@{\ }l@{\ }ccc@{\ }ccc@{\ }ccc@{\ }ccc@{\ }ccc}
        \toprule
         \multirow{2}[2]{*}{\textbf{Type}} & \multirow{2}[2]{*}{\textbf{Method}}
         & \multicolumn{3}{c}{\textbf{Book}}
         & \multicolumn{3}{c}{\textbf{Classroom}}
         & \multicolumn{3}{c}{\textbf{Monk}}
         & \multicolumn{3}{c}{\textbf{Room}} 
         & \multicolumn{3}{c}{\textbf{Kitchen}} \\
         \cmidrule(rl){3-5} \cmidrule(rl){6-8} \cmidrule(rl){9-11} \cmidrule(rl){12-14} \cmidrule(rl){15-17}
         && \textbf{PSNR}$\uparrow$ & \textbf{SSIM}$\uparrow$ & \textbf{LPIPS}$\downarrow$ 
         & \textbf{PSNR}$\uparrow$ & \textbf{SSIM}$\uparrow$ & \textbf{LPIPS}$\downarrow$ 
         & \textbf{PSNR}$\uparrow$ & \textbf{SSIM}$\uparrow$ & \textbf{LPIPS}$\downarrow$ 
         & \textbf{PSNR}$\uparrow$ & \textbf{SSIM}$\uparrow$ & \textbf{LPIPS}$\downarrow$ 
         & \textbf{PSNR}$\uparrow$ & \textbf{SSIM}$\uparrow$ & \textbf{LPIPS}$\downarrow$ \\
        \midrule
         $\mathcal{S}$ & Baseline 
         & 22.53 & 0.796 & 0.293
         & 28.71 & 0.902 & 0.261
         & 27.15 & 0.848 & 0.281
         & 30.70 & 0.912 & 0.183
         & 33.43 & 0.957 & 0.138
         \\
         \cmidrule{1-17}
         \multirow{4}{*}{$\mathcal{V}$} & Baseline
         & 11.92 & 0.454 & 0.597
         & 12.83 & 0.542 & 0.660
         & 15.81 & 0.535 & 0.542
         & 13.28 & 0.599 & 0.643
         & 18.24 & 0.718 & 0.496
        \\
         & ADOP
         & 22.15 & 0.824 & \textbf{\rred{0.291}}
         & 21.04 & 0.800 & 0.345 
         & 21.92 & 0.764 & \textbf{\blue{0.392}}
         & 19.25 & 0.834 & 0.329 
         & 20.13 & 0.827 & 0.280
        \\
         & NeRF-A
         & \textbf{\rred{28.44}} & \textbf{\rred{0.873}} & 0.310
         & \textbf{\blue{29.30}} & \textbf{\blue{0.895}} & \textbf{\blue{0.295}}
         & \textbf{\blue{27.33}} & \textbf{\blue{0.793}} & 0.398 
         & \textbf{\rred{30.32}} & \textbf{\blue{0.891}} & \textbf{\rred{0.234}}
         & \textbf{\blue{31.30}} & \textbf{\blue{0.928}} & \textbf{\blue{0.233}}
         \\
         & Ours
         & \textbf{\blue{27.49}} & \textbf{\blue{0.837}} & \textbf{\blue{0.292}} 
         & \textbf{\rred{29.87}} & \textbf{\rred{0.908}} & \textbf{\rred{0.284}} 
         & \textbf{\rred{28.27}} & \textbf{\rred{0.852}} & \textbf{\rred{0.297}} 
         & \textbf{\blue{28.70}} & \textbf{\rred{0.900}} & \textbf{\blue{0.291}} 
         & \textbf{\rred{31.53}} & \textbf{\rred{0.936}} & \textbf{\rred{0.156}} \\
         
        
         \bottomrule
        \end{tabular}
    }
    \label{table:quan_syn}
\end{table}

\paragraph{Comparison}
Our HDR-Plenoxels learn HDR radiance fields from LDR input images with various camera conditions at real scenes.
The results in \Fref{fig:qual_histogram} represent the novel view synthesis of HDR images and tone mapping from HDR to LDR images in real datasets.
Compared to previous work focusing on single perspective HDR~\cite{debevec_1996_modeling}, our method can reconstruct HDR radiance fields from varying camera LDR images with several multi-view. We can render novel LDR views from reconstructed HDR radiance fields with an explicitly controllable tone mapping module.

\begin{wrapfigure}{r}{0.50\linewidth}
\setlength{\belowcaptionskip}{-1mm}
  \centering
  \includegraphics[width=1.0\linewidth]{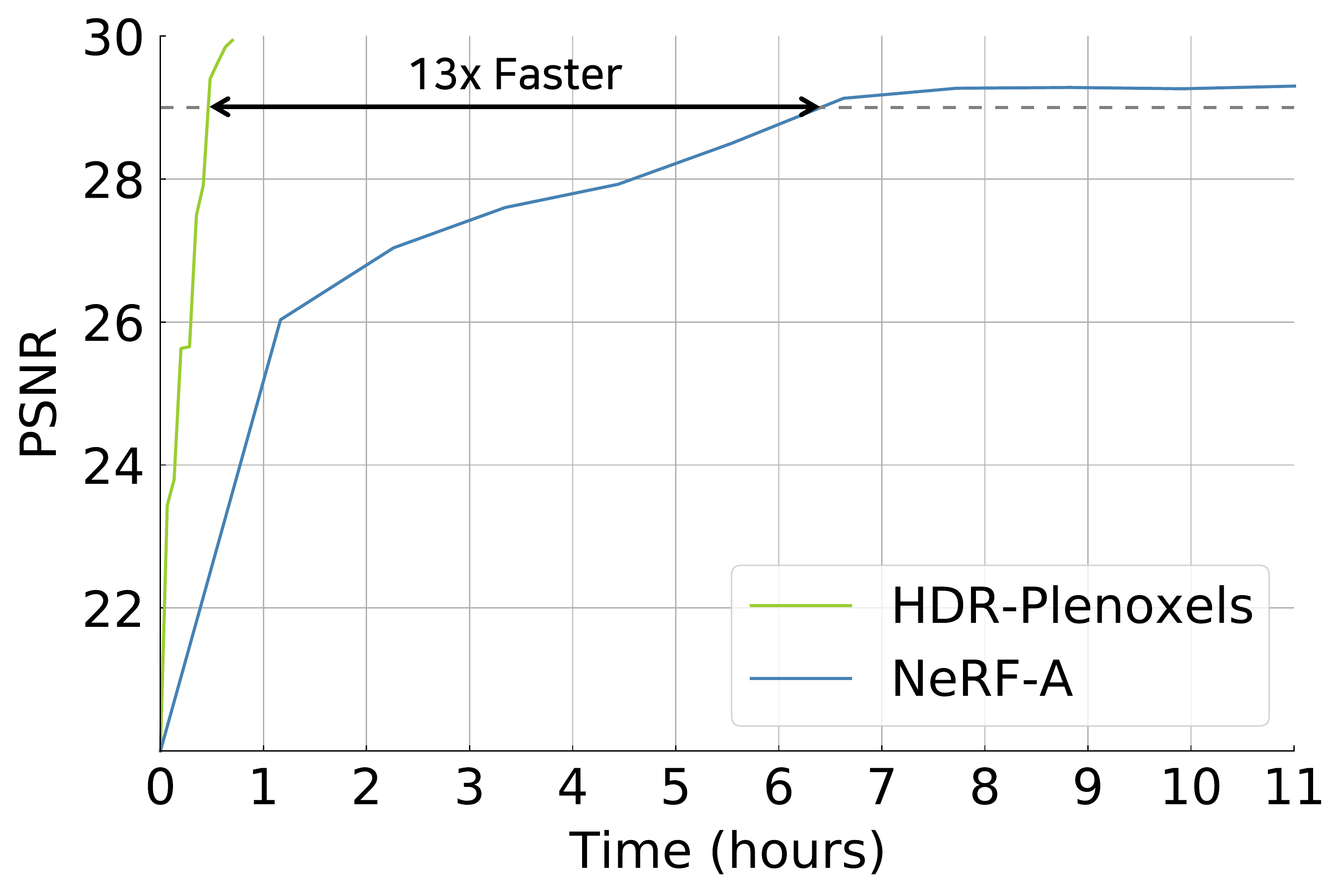}
  \caption{\textbf{Computation time.}}
  \label{fig:compuation_time_nerf_a}
\end{wrapfigure}
Quantitative results are summarized in \Tref{table:quan_syn}.
NeRF-A~\cite{martin-brualla_nerf_w_2021} shows comparable novel view synthesis results.
However, it cannot explicitly decompose each 
camera condition, such as exposure, white balance, and CRF, because 
that information is implicitly entangled in the embedding;
thus, it cannot predict \emph{radiance}
in contrast to 
ours.
\ug{NeRF-A also needs considerable training time compared to our neural networks free method as shown in \Fref{fig:compuation_time_nerf_a}.}
To reach PSNR 29, our method takes 30 min, but NeRF-A takes 6 h and 30 min, which is 13 times larger.
\ug{We use RTX 3090 for training.}
\begin{figure}[t]
    \setlength{\belowcaptionskip}{-1mm}
    \centering
    \includegraphics[width=1\linewidth]{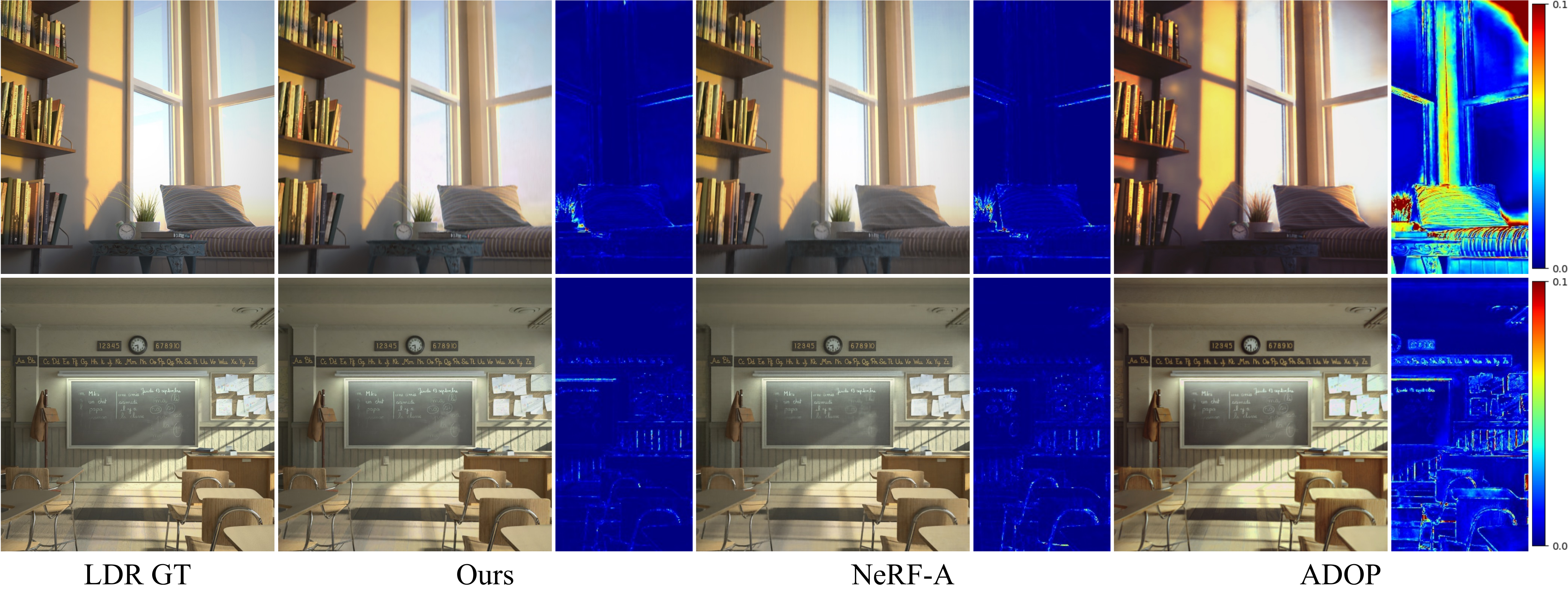}
    \caption{\textbf{Comparison in synthetic scenes with varying camera conditions.
    }
    The left half of an image is used for training the tone mapping module $\mathcal{T}$, and the right half is for the test.
    By applying the trained $\mathcal{T}$ at left half, we can synthesize the novel view images.
    MSE maps of each result are on the right of the corresponding render results.}
    \label{fig:qual_baseline}
\end{figure}

\begin{figure}[t]
    \centering
    \includegraphics[width=1\linewidth]{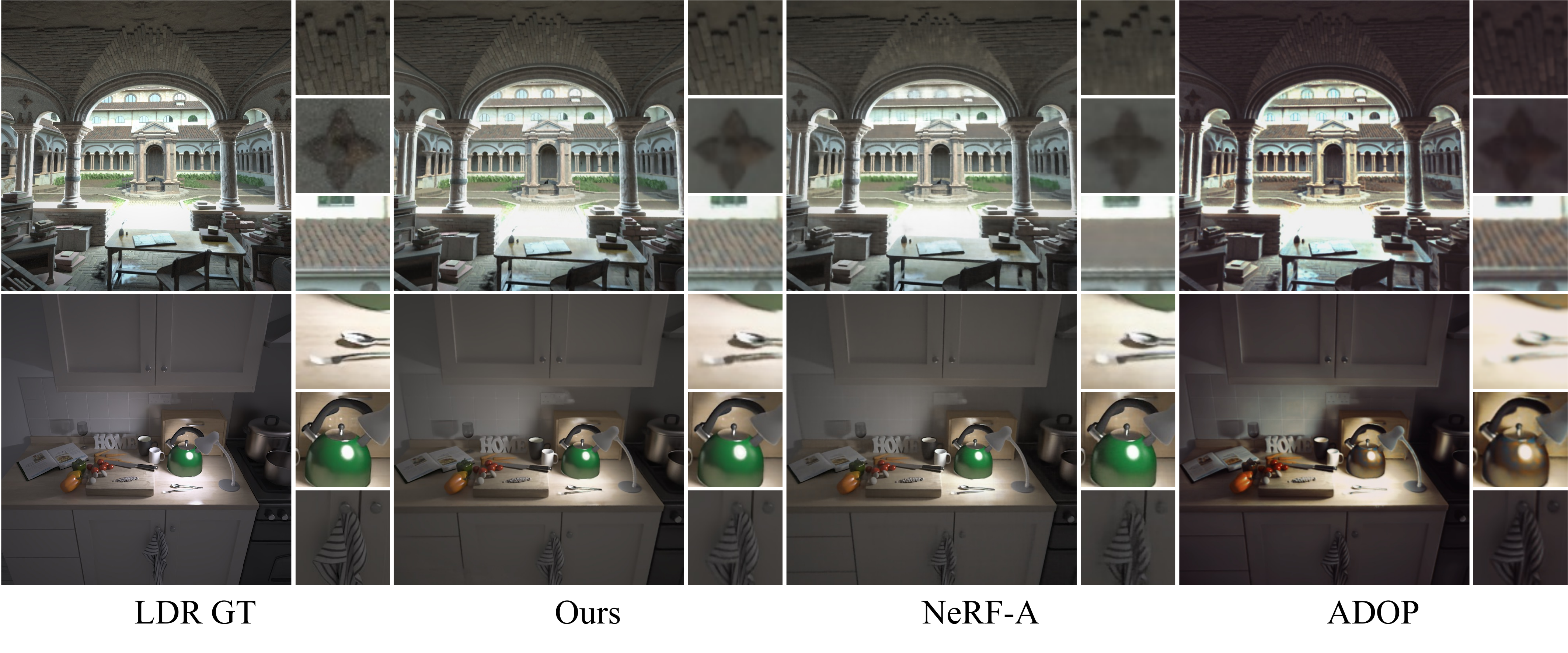}
    \caption{\textbf{Qualitative results from the experiments between other counterparts.}
     Our result represents 
    \ug{fine-grained and proper color rendering results.}
     All experiments are trained with various \ug{radiometric} conditions. The left half of an image is used in training, and the right half is only used to measure quantitative results.
    }
    \label{fig:qual_detail}
\end{figure}

The qualitative results of rendered novel LDR views on our synthetic dataset are shown in~\Fref{fig:qual_baseline} and~\Fref{fig:qual_detail}. 
ADOP tends to incorrectly estimate the camera components such as white balance, vignetting, and CRF, leading to inaccurate color-mapped rendering results.
\moon{The rendering result of NeRF-A shows comparable quality to ours overall, but the details tend to be deficient.}
On the contrary,
our HDR-Plenoxels can render relatively accurate LDR images in both aspects of color and geometry.


ADOP~\cite{ruckert_adop_2021} shows deficient novel view rendering results compared to our method.
In \Fref{fig:qual_detail}, they cannot reconstruct fine-grained geometry structures compared to ours, and color information is also ambiguous.
The error map results of ours also show better than ADOP results as shown in \Fref{fig:qual_baseline}.
Therefore, the visual quality of the novel view results of ADOP is not satisfying, and we outperform quantitative results with PSNR metrics at \Tref{table:quan_syn}.

\begin{figure}[t]
    \centering
    \includegraphics[width=1\linewidth]{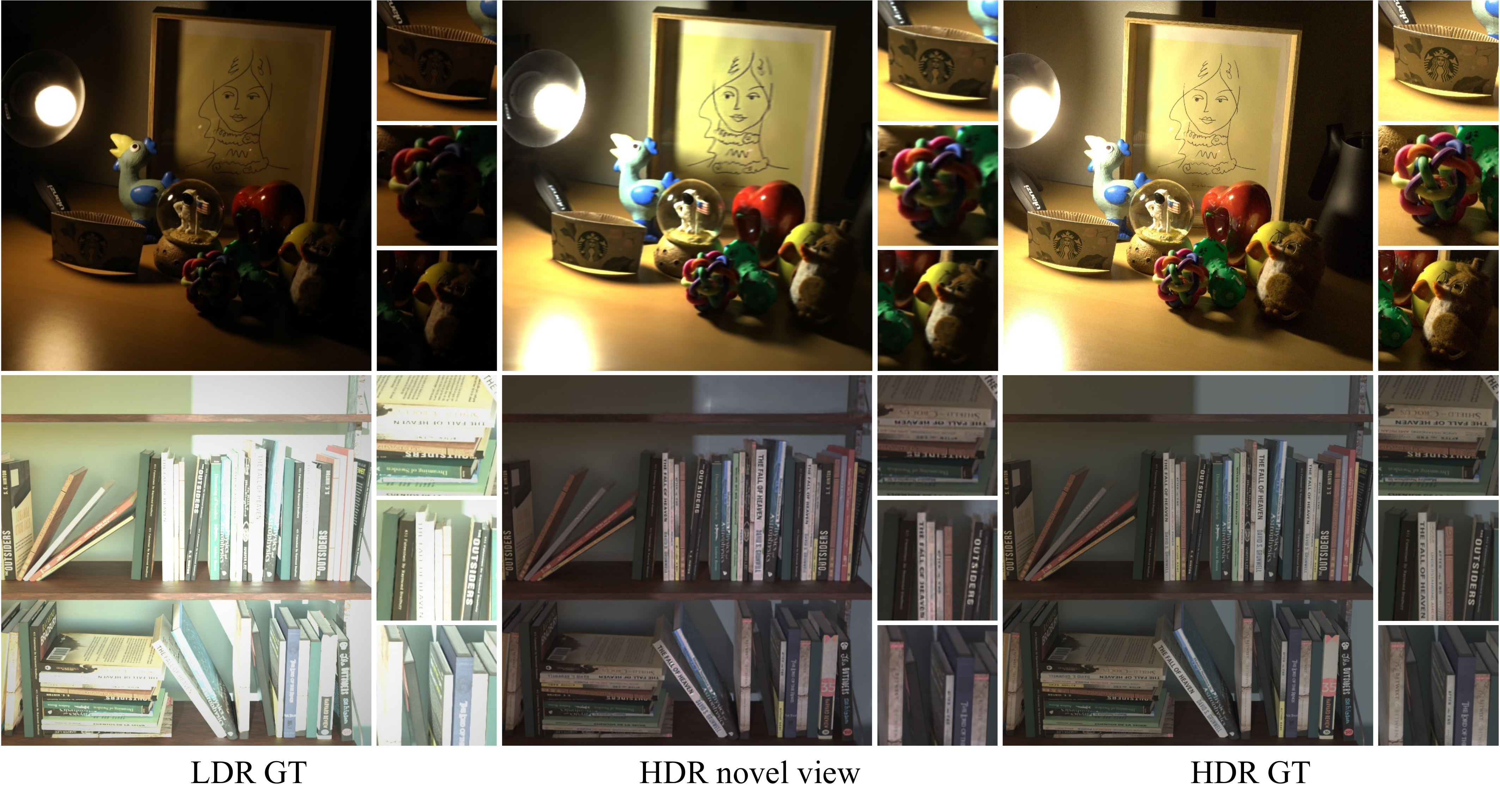}
    \caption{\textbf{High dynamic range rendering at saturation points.}
    \ug{Our novel view rendering results show robustness to under-saturation ($1^{\mathrm{st}}$ row) and over-saturation ($2^{\mathrm{nd}}$ row) points because we train the HDR radiance fields.}
    }
    \label{fig:qual_saturation}
\end{figure}

To verify that our model can restore HDR radiance field robust to under- or over- saturated points, we qualitatively compared HDR details on saturation points as shown in \Fref{fig:qual_saturation}.
There are severe dark or bright points, so some regions are saturated and cannot distinguish color or geometry. 
\ug{We train with these saturated LDR input images at our HDR-Plenoxels. They can render high-quality HDR novel views, which means we can handle wide dynamic ranges and render non-saturated novel HDR views.}

We also compare our HDR-Plenoxels and original Plenoxels~\cite{yu_plenoxels_2021} with images of different camera settings in \Fref{fig:qual_in_ours}.
The novel view synthesis results of the Plenoxels with static camera condition are considered as our upper bound performance and represent fine-detailed rendering results with clear color estimation.
With the varying camera setting, the Plenoxels fail to reconstruct proper 3D geometry, especially in the right-half of images unseen during training.
These results imply that previous volume rendering methods such as \kjs{the} Plenoxels are prone to degrade at varying camera conditions due to their static \ugkim{photometric} assumption, which needs to be compensated with additional regularization.
With our tone mapping module, the quality of novel view synthesis improves considerably by showing comparable details to original Plenoxels trained with static LDR images.
This demonstrates that our tone mapping module precisely disentangles varying cameras and properly reconstructs informative HDR radiance fields.

\begin{wraptable}{r}{0.52\linewidth}
    \setlength{\belowcaptionskip}{0mm}
    \centering
    \caption{
    \textbf{Effect of tone mapping components on novel view synthesis.}
    \textbf{WB} stands for white balance, \textbf{VIG} for vignetting, and \textbf{CRF} for camera response function.
    (A) is the baseline. 
    (D) is the performance of ours.
    }
    \resizebox{1.0\linewidth}{!}{ 
        \begin{tabular}{l@{\quad}ccc@{\quad}ccc}
        \toprule
         & \textbf{WB} & \textbf{VIG} & \textbf{CRF} & \textbf{PSNR}$\uparrow$ & \textbf{SSIM}$\uparrow$ & \textbf{LPIPS}$\downarrow$ \\
        \midrule 
        (A) &  & &
            & $14.42$ & $0.569$ & $0.587$ \\
        (B) & $\checkmark$ & &
            & $23.03$ & $0.811$ & $0.355$ \\
        (C) & $\checkmark$ & $\checkmark$ & 
            & $21.12$ & $0.799$ & $0.352$ \\
        (D) & $\checkmark$ &  & $\checkmark$ 
            & \textbf{29.34} & \textbf{0.876} & \textbf{0.294} \\
        (E) & $\checkmark$ & $\checkmark$ & $\checkmark$ 
            & $26.73$ & $0.878$ & $0.264$ \\
        
        \bottomrule
        \end{tabular}
    }
    \label{table:ablation_2}
\end{wraptable}

\paragraph{Ablation Studies}
\moon{To verify the effectiveness of our tone mapping module components, we conduct an ablation study by removing each component as in \Tref{table:ablation_2}.
We report the averaged results of five synthetic datasets.
Comparing (A) with (B), the considerable performance improvement by the white balance module means that disentanglement of exposure and white balance helps to learn accurate geometry and color.
The performance degradation from (B) to (C) and from (D) to (E) implies that decomposing the vignetting effect of images is inconducive.}
According to the characteristics of our tone mapping module, which trains quickly according to the learning speed of SH, the vignetting function significantly increases the complexity of the model and hinders training.
\moon{The result in (D) demonstrates that all components of our tone mapping module are essential.}

\begin{wrapfigure}{r}{0.39\linewidth}
  \setlength{\abovecaptionskip}{-1mm}
  \setlength{\belowcaptionskip}{-1mm} 
  \centering
  \includegraphics[width=1.0\linewidth]{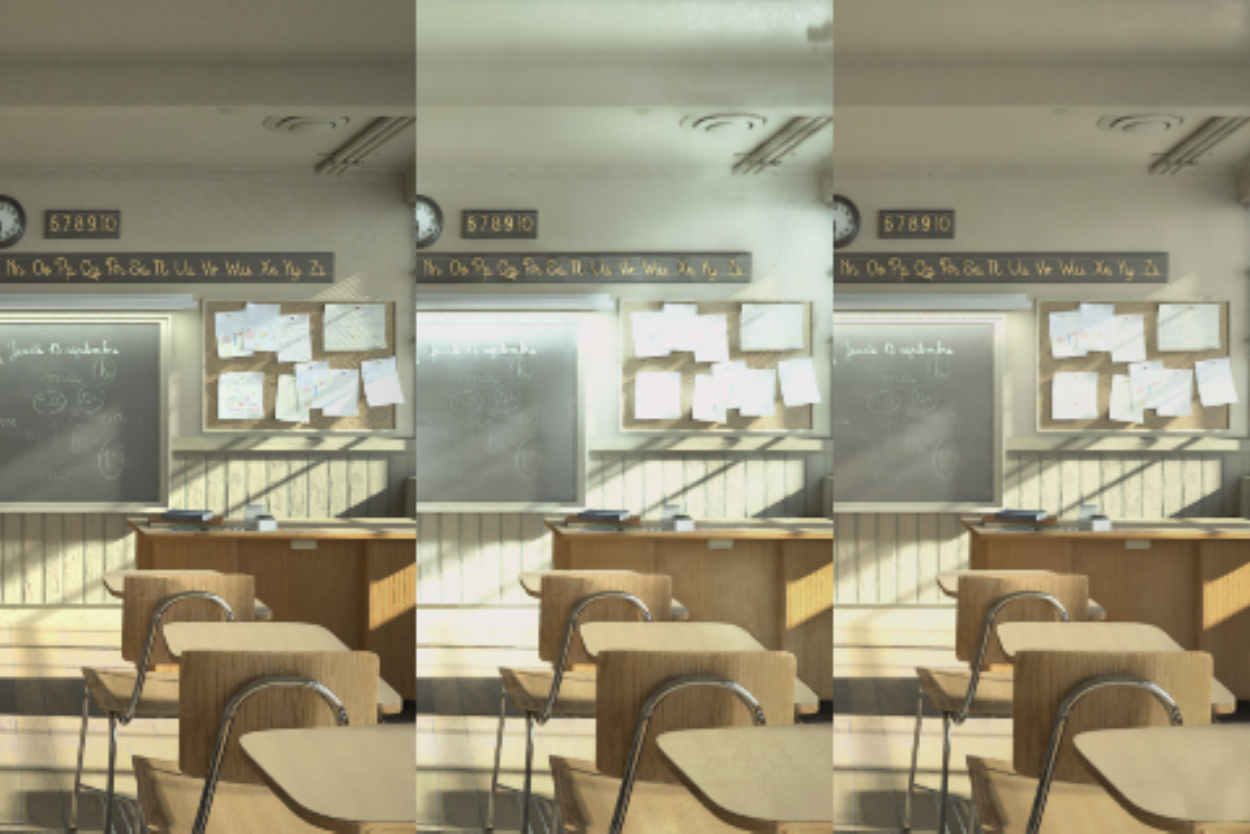}
  \caption{\textbf{Implicit CRF.} GT (left), MLP (middle), and ours (right). }
  \label{qual_crf}
\end{wrapfigure}
To confirm whether the CRF expression is suitable for defining camera non-linearity, we proceed an ablation study by replacing the CRF function.
Our method 
adopts a piece-wise linear \kjs{CRF} model 
that considers the mapping relationship of the pixel value.
NeRF-W and HDR-NeRF~\cite{huang_hdr-nerf_2021} suggest that
an implicit function can represent CRF.
We compare the performance between our explicit piece-wise linear function and implicit function such as multi-layer perceptrons (MLPs).
Following the concurrent work, HDR-NeRF, we replace our CRF with three MLPs to predict each color channel.
\ug{As shown in ~\Fref{qual_crf}, rendering results are degraded with MLP-based CRF.
The averaged PSNR of all synthetic datasets with ours is 28.18, and the MLP-based method is 26.11.}
We verify that our physically-based explicit tone mapping module
\ug{outperforms than MLP-based method}
since our CRF satisfies real-world physical conditions and can disentangle non-linear components correctly.

\section{Conclusion}
We present an HDR-Plenoxels, which learn to synthesize 3D HDR radiance field from multi-view LDR images of the varying camera by self-calibrating radiometric characteristics.
Distinctive from conventional HDR reconstruction methods, ours can get a novel view, depth, and 3D HDR radiance fields simultaneously in an end-to-end manner.
We investigate that the white balance and CRF functions are critical factors among the in-camera components and find an effective representation of the CRF function.
With these observations, we present a simple and straightforward physical-based tone mapping module, which can be easily attached to various volume rendering models extending one's usability.
\ug{Using the tone mapping module, we can take fine-grained HDR rendering results as well as LDR images of varying cameras, \eg, exposure, white balance, and CRF.}
The HDR radiance fields represent real-world scenes more realistically with a wide dynamic range similar to the way human sees. 
Our work could improve the experiences of many HDR-based applications, such as movie post-production.
Since we focus on covering HDR radiance fields of static scenes, it has room to improve in dealing with dynamic objects as future work.

\vfill
{\scriptsize\paragraph{Acknowledgment}
This work was partly supported by the National Research Foundation of Korea (NRF) grant funded by the Korea government (MSIT) (No. NRF-2021R1C1C1006799), Institute of Information \& communications Technology Planning \& Evaluation (IITP) grant funded by the Korea government(MSIT) (No.2022-0-00290, Visual Intelligence for Space-Time Understanding and Generation based on Multi-layered Visual Common Sense; and No.2019-0-01906, Artificial Intelligence Graduate School Program(POSTECH)).\par}

\bibliographystyle{splncs04}
\bibliography{egbib}


\clearpage

\renewcommand{\thefigure}{S\arabic{figure}}
\renewcommand{\thetable}{S\arabic{table}}
\setlength{\abovecaptionskip}{-2mm}
\setlength{\belowcaptionskip}{-2mm}
\setlength\intextsep{2mm}

\renewcommand{\thefootnote}{\fnsymbol{footnote}}



\pagestyle{headings}
\def\ECCVSubNumber{3703}  

\title{Supplementary Material for\\ ``HDR-Plenoxels: Self-Calibrating\\High Dynamic Range Radiance Fields''} 


\titlerunning{HDR-Plenoxels}

\author{Kim Jun-Seong\inst{1}\thanks{Authors contributed equally to this work.} \quad
Kim Yu-Ji\inst{2\ast} \quad
Moon Ye-Bin\inst{1} \quad
Tae-Hyun Oh\inst{1, 2}\thanks{Joint affiliated with Yonsei University, Korea.}}

\authorrunning{K. Jun-Seong et al.}

\institute{
${}^1$Dept. of Elect. Eng. \qquad ${}^2$Grad. School of AI
\\Pohang University of Science and Technology (POSTECH)\\
\email{\{junseong.kim, ugkim, ybmoon, taehyun\}@postech.ac.kr}\\
\url{https://github.com/postech-ami/HDR-Plenoxels}
}

\maketitle
%
\hypersetup{linkcolor=black}
\section*{Contents}

\noindent\hyperref[sec:A]{\textbf{A \ \ \ Technical Details}}\\
\hyperref[sec:A.1]{\text{\qquad A.1 \ \ \ Experimental Settings}} \\ 
\hyperref[sec:A.2]{\text{\qquad A.2 \ \ \ Counterpart Method Details}} \\ \\
\hyperref[sec:B]{\textbf{B \ \ \ Novel View Synthesis of Real Scenes}} \\
\hyperref[sec:B.1]{\text{\qquad B.1 \ \ \ Qualitative Results}} \\ 
\hyperref[sec:B.2]{\text{\qquad B.2 \ \ \ Quantitative Results}} \\ \\
\hyperref[sec:C]{\textbf{C \ \ \ Controllable Rendering at Novel View Synthesis}} \\ 
\hyperref[sec:C.1]{\text{\qquad C.1 \ \ \ Exposure and White Balance}} \\ 
\hyperref[sec:C.2]{\text{\qquad C.2 \ \ \ Camera Response Function}} \\
\hyperref[sec:C.3]{\text{\qquad C.3 \ \ \ Comparison between HDR-Plenoxels and NeRF in the Wild}} \\ \\
\hyperref[sec:D]{\textbf{D \ \ \ Additional Experiments }} \\ 
\hyperref[sec:D.1]{\text{\qquad D.1 \ \ \ Denoising Effects}} \\ 
\hyperref[sec:D.2]{\text{\qquad D.2 \ \ \ Extreme Camera Conditions}} \\ 
\hyperref[sec:D.3]{\text{\qquad D.3 \ \ \ Generality of the Tone Mapping Module}} \\ 

\noindent\rule{\linewidth}{0.2pt}
\hypersetup{linkcolor=blue}

\section*{Supplementary Material}
This supplementary material presents technical details, results, and experiments not included in the main paper due to the space limit.

\appendix

\section{Technical Details}\label{sec:A}
In this section, we explain the details of experimental settings (in \Sref{sec:A.1}) and method (in \Sref{sec:A.2}) of
our high dynamic range radiance Plenoxels (HDR-Plenoxels) and other counterparts.
We build our code based on PyTorch open library~\cite{NEURIPS2019_9015} and use one NVIDIA GeForce RTX 3090 GPU or A100 for training and rendering novel view synthesis.

\subsection{Experimental Settings}\label{Sec:A.1}

\paragraph{Synthetic Dataset}
To generate the synthetic dataset, we use Blender~\cite{blender} to modify the various camera.
Each image is created in OpenEXR format and conducted post-processing like changing exposure and white balance at high dynamic range (HDR) colorspace for comparing physical camera tone mapping.
Synthetic scenes have five different scenes, \ie, 
book~\cite{book_room}, classroom (CC0)~\cite{blender_classroom}, kitchen (CC-BY)~\cite{kitchen}, palace (CC-BY)~\cite{blender_monk}, and room~\cite{book_room}.
To demonstrate the ability of novel view synthesis to our HDR-Plenoxels, we generate synthetic data with complex geometry and various radiometric conditions.
In the test stage, we split the test image into left and right half.
We train the left half of a test image and test with the right half because our method needs trained parameters of the tone mapping module for rendering.
The left half of the test image is trained, and the right half is used at test time.
Exposure was set to $\pm$3EV at the basis image, white balance was applied by multiplying each color channel with 1.25 separately, and camera response function (CRF) was set to gamma correction with $\gamma=3$.
We show experimental results on five synthetic datasets. 
Each scene is created at an 800 $\times$ 800 pixels resolution, with views sampled from the roughly forward-facing camera. We use \ug{43} views of each scene as training input and \ug{7} for testing.

\paragraph{Real Dataset}
We take all real scenes with exposure bracketing setting and changing white balance measured in Kelvin.
Various camera shooting conditions, \ie,
exposure value with three intervals and white balance with 3000K, 3500K, and 4000K, are applied sparsely to whole datasets.
Our real datasets are taken with Canon EOS 5D Mark $\uppercase\expandafter{\romannumeral4}$, captured 30 to 50 views, and taken 1/5 as a test set. 
Most images are taken in strong sunlight or darkroom, so each pixel is easily saturated and needs an HDR colorspace to represent the accurate color of a pixel. 
The experimental results on four real datasets are shown in Sec.~\ref{sec:B}.

\begin{figure}[t]
    \centering
    \includegraphics[width=1\linewidth]{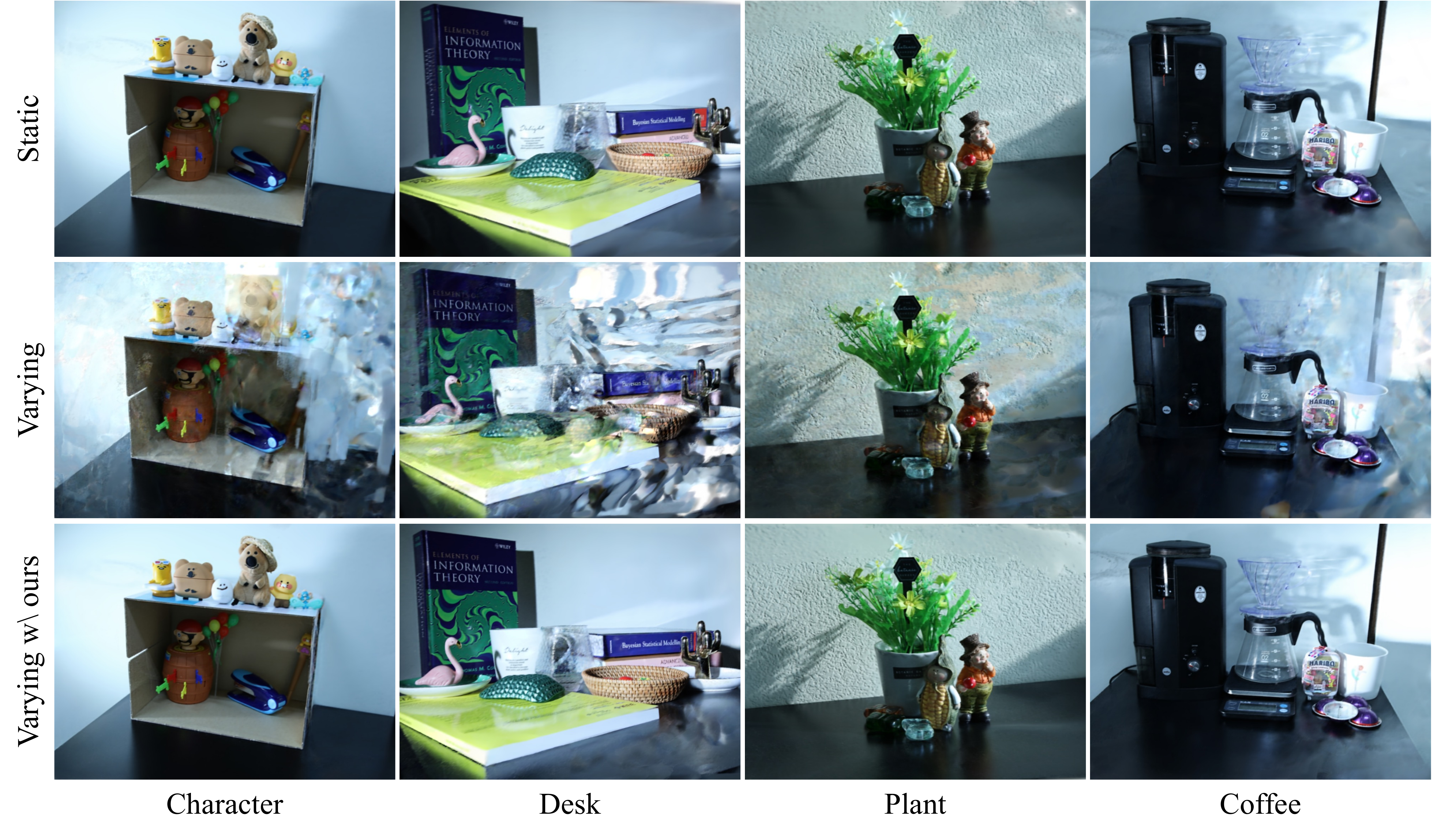}
    \caption{\textbf{Results of static and varying camera settings in real scenes.}
    The static and varying represent different camera conditions (\ie, exposure, white balance, and camera response function (CRF)).
    In static camera conditions, all views of the scene have the same exposure, white balance, and CRF, and the varying one is vice versa.
    The first two rows are results from original Plenoxels, and the last row is from HDR-Plenoxels.
    Each column represents a different real scene.}
    \label{sup_fig:real_in_ours}
\end{figure}
\paragraph{Hyperparameters}
We follow the learning processes suggested by Plenoxels~\cite{yu_plenoxels_2021} with some modifications.
We train our HDR-Plenoxels 
with 10 epochs, a total of 128,000 iterations each, with 5,000 rays per batch.
We set a learning rate of the spherical harmonics (SH) and tone mapping parts to pure exponential decay.
The learning rate of SH and tone mapping starts from 0.01, and both decays to $5\cdot10^{-5}$ at step 250,000 to match each other's learning speed. 
Voxel opacity $\sigma$ is updated using a delayed exponential function with decaying up to 0.05 during 250,000 iterations.

For learning stability of opacity $\sigma$ and SH, total variance (TV) loss is applied only for the $3$ epochs until the first upsampling process.
We apply the weight of TV loss for opacity $\sigma$ and SH as $\lambda_\textrm{TV,$\sigma$}=5\cdot 10^{-4}$, $\lambda_\textrm{TV,SH}=1\cdot10^{-2}$.
Each loss is updated with an RMSProp optimizer.

Our HDR-Plenoxels are a voxel grid-based method, so it is important to find the proper range of the initial grid for retaining the expressible volume.
Several scenes contain large depth in synthetic Blender data, which is hard to express with a default concentric sphere grid.
To properly determine the grid range, we first compute the rough 3D geometry of the scene, and we then obtain the camera poses with the 3D boundary of the scene through the COLMAP~\cite{schoenberger2016mvs, schoenberger2016sfm} sparse reconstruction.
We start grid voxel resolution in (128, 128, 64)
and then upsample the resolution in the order of (256, 256, 128), (512, 512, 256), and (800, 800, 512) after each 25,600 iterations.

We similarly obtain 
the unknown camera pose 
through COLMAP and initialize the grid in the real data.
In particular, 
we conduct undistortion of the entire image through calibration before the training learning due to the lens distortion frequently appearing in the image.
The resolution of the real image is updated in the order of (128, 128, 128), (256, 256, 256), and (750, 600, 300).\label{sec:A.1}
\begin{figure}[t]
\setlength{\abovecaptionskip}{0mm} 
    \centering
    \includegraphics[width=1\linewidth]{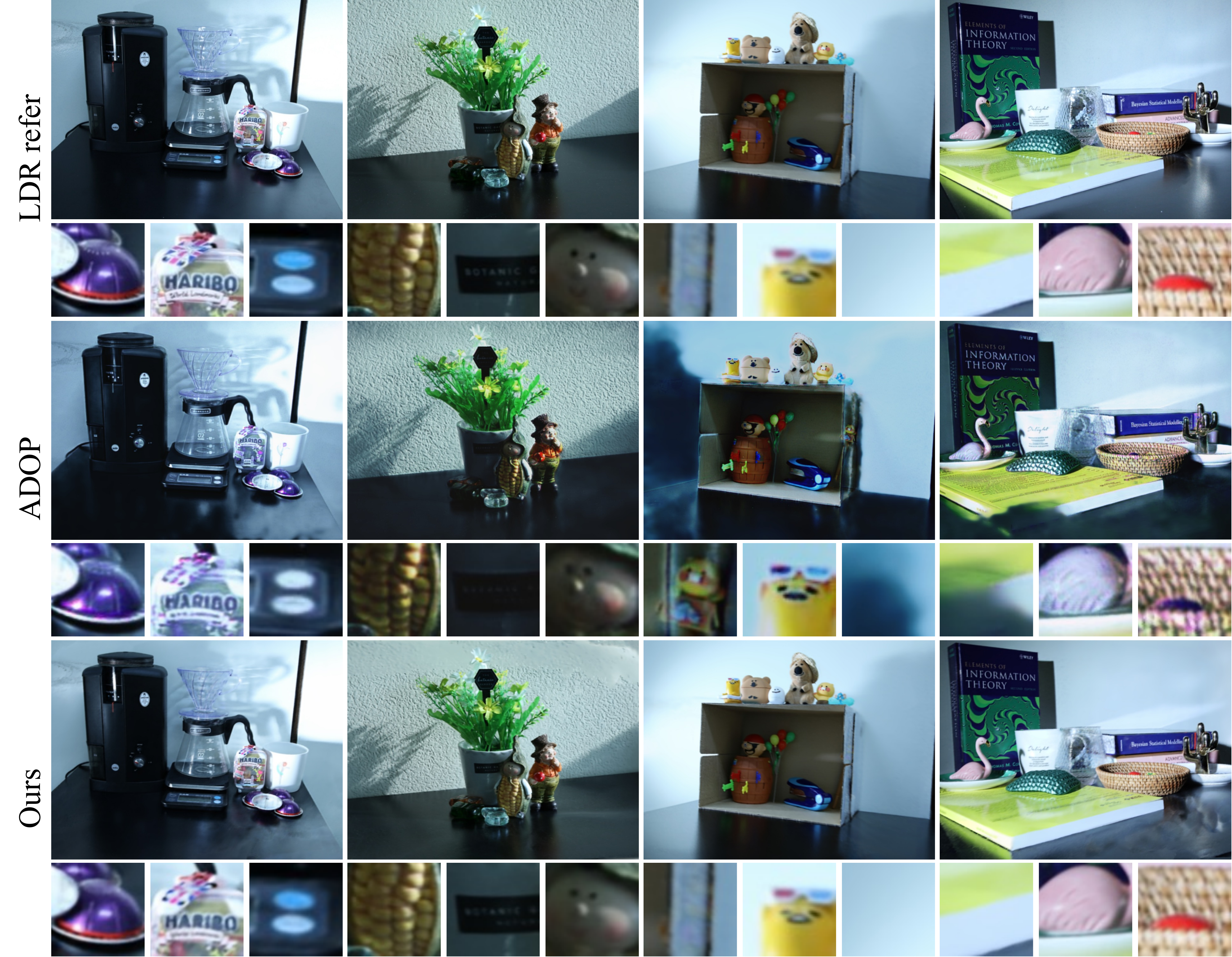}
    \caption{\textbf{Comparisons of qualitative results to a counterpart.}
    The first row represents the LDR reference image, which is used at training.
    All experiments are trained with varying conditioned data and rendered with the tone mapping stage.
    Our results represent fine-grained rendering results compared to the ADOP.
    Especially, ours shows satisfying color representations compared to ADOP.
    }
    \label{sup_fig:baseline}
\end{figure}
\subsection{Counterpart Method Details}\label{Sec:A.2}

\paragraph{NeRF in the Wild (NeRF-W)~\cite{martin-brualla_nerf_w_2021}}
We conduct the comparison of NeRF-W based on the PyTorch version of NeRF-W implementation~\cite{nerf_w_github} with the same width and depth of the original model~\cite{martin-brualla_nerf_w_2021}.
We train the model with the left half side of the images in the training set and test the novel view synthesis on the right half side of the images.
We use NeRF-A (appearance) without transient embeddings for a fair comparison because our dataset has no transient parts.

\paragraph{Approximate Differentiable One-Pixel Point Rendering (ADOP)~\cite{ruckert_adop_2021}}
We use the official code of~\cite{adop_github}.
Given the COLMAP dense reconstruction results, we train the ADOP model for 100 epochs, which are enough to show the model performance on the novel view synthesis.
We use the mask covering the right half side of the images for dense reconstruction and model optimization, following NeRF-W~\cite{martin-brualla_nerf_w_2021}.

\label{sec:A.2}
\section{Novel View Synthesis of Real Scenes}\label{sec:B}
To evaluate our method, we compare it to the novel view synthesis counterparts, which handle the varying appearance of images.
We present several experimental results to verify the effectiveness of our method in qualitative and quantitative views of real scenes.

\subsection{Qualitative Results}
We compare our HDR-Plenoxels with original Plenoxels in different camera settings, and its qualitative results are in \Fref{sup_fig:real_in_ours}.
The results of original Plenoxels with static camera condition located in the first row, represent our upper bound performance of novel view synthesis.
As described in (\Sref{sec:A.2}), the right half of the test image is unseen data, meaning novel view synthesis.
In the second row, original Plenoxels with varying camera settings show poor renderings results, especially in the right half, where we split as test images.

\begin{table}[t]
\setlength{\belowcaptionskip}{1mm}
    \centering
     \caption{\textbf{Quantitative results of novel view synthesis on real scenes.}
    $\mathcal{S}$ denotes the static and $\mathcal{V}$ is the varying datasets.
    The blue and red colors stand for the \textbf{\rred{best}} and the \textbf{\blue{second best}}, respectively.
    We report the averaged results of all the views in each test data. 
    Our method shows the highest or the second-best performance compared to other models.
    }
    \resizebox{\textwidth}{!}{ 
        \begin{tabular}{c@{\ }l@{\ }ccc@{\ }ccc@{\ }ccc@{\ }ccc}
        \toprule
         \multirow{2}[2]{*}{\textbf{Type}} & \multirow{2}[2]{*}{\textbf{Method}}
         & \multicolumn{3}{c}{\textbf{Character}}
         & \multicolumn{3}{c}{\textbf{Desk}}
         & \multicolumn{3}{c}{\textbf{Plant}} 
         & \multicolumn{3}{c}{\textbf{Coffee}} \\
         \cmidrule(rl){3-5} \cmidrule(rl){6-8} \cmidrule(rl){9-11} \cmidrule(rl){12-14} 
         &
         & \textbf{PSNR}$\uparrow$ & \textbf{SSIM}$\uparrow$ & \textbf{LPIPS}$\downarrow$ 
         & \textbf{PSNR}$\uparrow$ & \textbf{SSIM}$\uparrow$ & \textbf{LPIPS}$\downarrow$ 
         & \textbf{PSNR}$\uparrow$ & \textbf{SSIM}$\uparrow$ & \textbf{LPIPS}$\downarrow$ 
         & \textbf{PSNR}$\uparrow$ & \textbf{SSIM}$\uparrow$ & \textbf{LPIPS}$\downarrow$ \\
        \midrule
         $\mathcal{S}$ & Baseline 
         & 32.40 & 0.955 & 0.278
         & 25.53 & 0.895 & 0.303
         & 24.58 & 0.833 & 0.324
         & 25.87 & 0.922 & 0.301
         \\
         \cmidrule{1-14}
         \multirow{3}{*}{$\mathcal{V}$} & Baseline
         & \textbf{\blue{19.13}} & 0.762 & 0.526
         & \textbf{\blue{13.75}} & \textbf{\blue{0.553}} & \textbf{\blue{0.518}}
         & \textbf{\blue{21.29}} & \textbf{\blue{0.623}} & 0.511
         & 17.49 & 0.751 & 0.476
        \\
         & ADOP
         & 17.56 & \textbf{\blue{0.801}} & \textbf{\rred{0.114}} 
         & 10.298 & 0.390 & \textbf{\blue{0.518}}
         & 18.26 & 0.529 & \textbf{\rred{0.192}} 
         & \textbf{\blue{18.44}} & \textbf{\blue{0.822}} & \textbf{\rred{0.085}}
         \\
         & Ours
         & \textbf{\rred{33.14}} & \textbf{\rred{0.960}} & \textbf{\blue{0.343}} 
         & \textbf{\rred{28.32}} & \textbf{\rred{0.907}} & \textbf{\rred{0.312}} 
         & \textbf{\rred{24.27}} & \textbf{\rred{0.790}} & \textbf{\blue{0.369}} 
         & \textbf{\rred{27.40}} & \textbf{\rred{0.928}} & \textbf{\blue{0.269}} 

         \\
        
         \bottomrule
        \end{tabular}
    }
    \label{sup_table:real_all}
\end{table}

We also compare the qualitative results with a counterpart model, ADOP, and our HDR-Plenoxels, as shown in \Fref{sup_fig:baseline}.
Both ADOP and our results represent comparable novel view synthesis with predicting fine-detailed 3D geometry.
However, ADOP shows biased results in estimating satisfying color and shadows compared to ours.
If a hole occurs during the point-cloud generation, the reconstruction result also shows vacancy in the rendered result because ADOP is a point-cloud-based rendering model.
The training stage of ADOP is unstable if they are in local optima, resulting in the imperfect color of novel view synthesis.
In contrast, our HDR-Plenoxels successfully reconstruct 3D real scenes with achieving a highly favorable tone-mapping stage.
Due to the properties of SH, which regularize complex color information on a few bases, we can optimize the model fast and stable.\label{sec:B.1}
\subsection{Quantitative Results}

We compare our HDR-Plenoxesl with other methods, consisting of the original Plenoxels (baseline) in both static and varying conditioned data and ADOP in the varying data, and its quantitative results are in \Tref{sup_table:real_all}. 


ADOP performs the gamma correction ($\gamma = 1/2.2$) as default by design, not the learned CRF function; thus, we first linearized the image by applying inverse gamma correction and then learned CRF.
ADOP is based on a dense reconstructed point cloud, which can recover structurally detailed scenes.
However, ADOP shows low performance in inferring the overall white balance and changing colors, including shading, which leads to low scores on SSIM and PSNR.
In contrast, our HDR-Plenoxels show overall high performance in all three metrics demonstrating that ours can understand physically appropriate tone-mapping parts and 3D structure as well.\label{sec:B.2}

\section{Controllable Rendering at Novel View Synthesis}\label{sec:C}
\begin{figure}[t]
\setlength{\abovecaptionskip}{-1mm}
    \centering
    \includegraphics[width=1\linewidth]{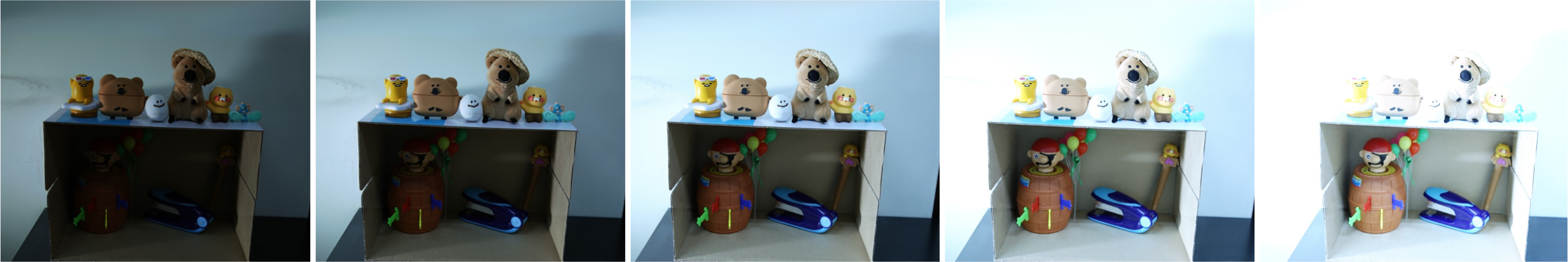}
    \caption{\textbf{Novel view synthesis results with varying exposure rendering.}
    The exposure condition changes from dark to bright from left to right.
    The novel view rendering result of the middle column has a basis exposure value.
    }
    \label{sup_fig:exp_change}
\end{figure}
\begin{figure}[t]
    \centering
    \includegraphics[width=1\linewidth]{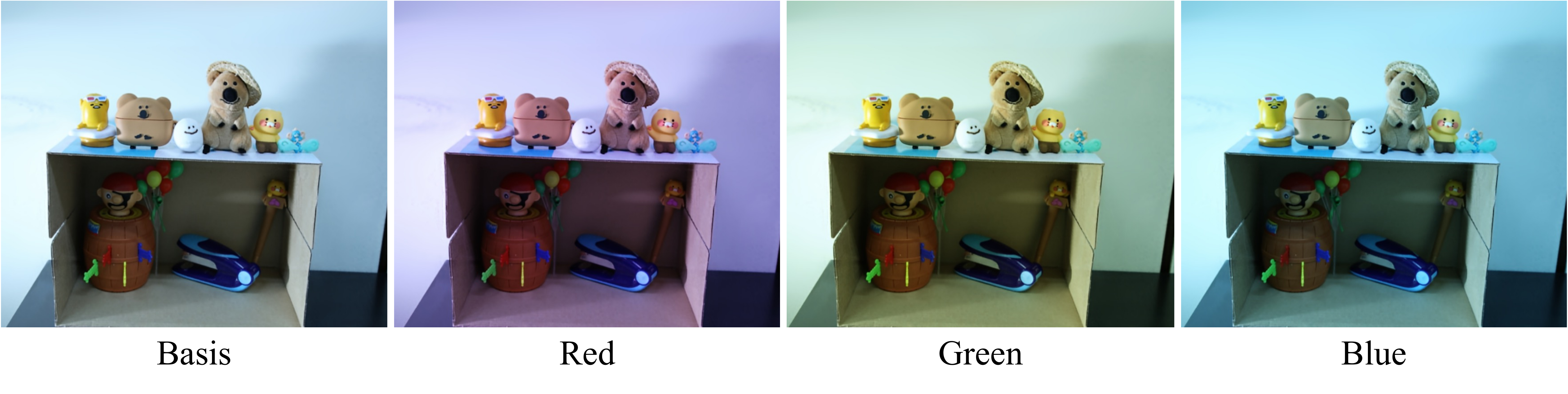}
    \caption{\textbf{Novel view synthesis results with varying white balance rendering.}
    The white balance changes \wrt red, green, and blue in the results of the second, third, and last columns, respectively.
    The novel view rendering result of the first column has a basis white balance value.
    }
    \label{sup_fig:wb_change}
\end{figure}

This section represents controllable rendering results with arbitrary exposure, white balance, and CRF settings.
Our tone mapping module consists of two stages, \ie, white balance and CRF.
Each stage is designed following the physical properties and represented by an explicit function to change the values of each stage.
To eliminate the ambiguity between exposure and white balance, we apply a white balance module with a scale suggested by Kim~\etal~\cite{kim_new_2012}.
In our white balance parameters, the exposure value is represented by a scale of white balance, which enables us to control exposure value as well.

\subsection{Exposure and White Balance}

We show controllable rendering results of arbitrary exposure in \Fref{sup_fig:exp_change}.
To change the exposure value, we set the basis exposure value by globally averaging the white balance values of full view.
With scaling basis exposure value, we can control the exposure of novel view rendering.

To control the white balance, we set the basis of white balance by channel-wise averaging the white balance values of full view.
By changing the respective red, green, and blue components on the basis of the white balance, we can control the white balance of novel view rendering, as shown in \Fref{sup_fig:wb_change}. 
This controllable rendering allows us to get the most advantages of synthesizing novel views with HDR, enabling editing HDR images and video in novel views through freely controllable rendering.\label{sec:C.1}
\subsection{Camera Response Function}
\begin{figure}[t]
    \centering
    \includegraphics[width=1\linewidth]{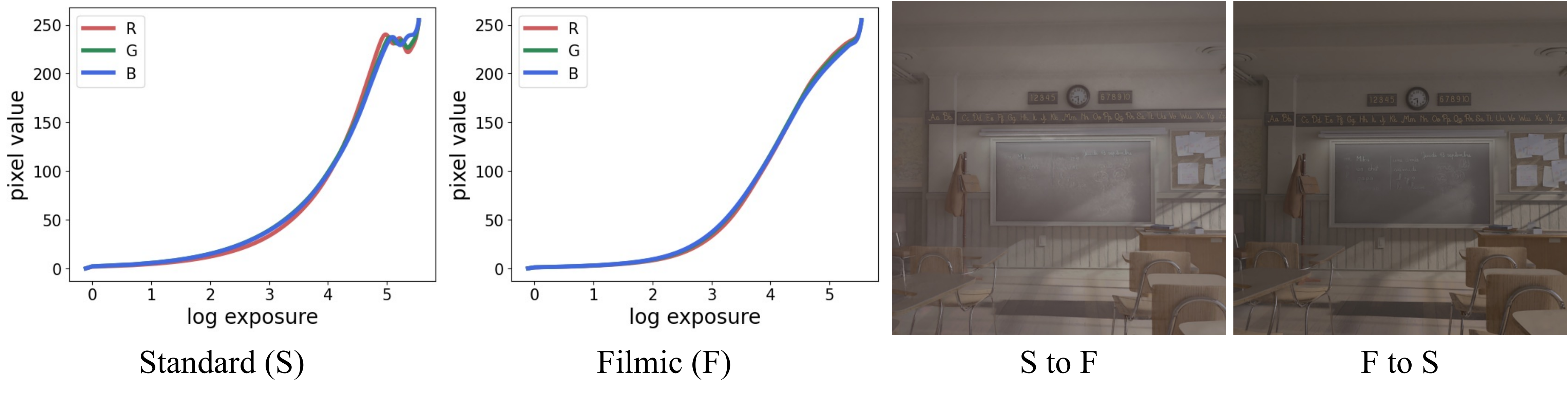}
    \caption{\textbf{Novel view synthesis changing with CRF.}
    The first two plots show CRF learned from images modified with two different view transforms implemented in Blender (Standard and Filmic).
    By exchanging learned CRFs, we can render the same synthetic scene in different styles.
    }
    \label{sup_fig:crf_change}
\end{figure}

    

To verify the ability of our CRF module, we train our HDR-Plenoxels with images of two different CRF rendering the same scene and transfer each CRF.
After transferring each CRF, the novel view rendering results show distinct rendering styles, as shown in \Fref{sup_fig:crf_change}.
The deliberate comparisons of CRF show the shape difference between filmic and standard CRF according to each RGB color channel.
The results of transferred CRF rendering imply that HDR-Prenoxels can learn robustly even in various CRFs. Also, we can apply diverse CRF to enable various rendering styles and more free HDR image and video editing.\label{sec:C.2}
\subsection{Comparison between HDR-Plenoxels and NeRF in the Wild}

\begin{wraptable}{r}{0.45\linewidth}
\setlength{\belowcaptionskip}{1mm}
    \centering
     \caption{\textbf{Controllable rendering comparison at a classroom image.}
    }
    \resizebox{0.95\linewidth}{!}{ 
        \begin{tabular}{ccc@{\ }ccc}
        \toprule
         \multicolumn{3}{c}{\textbf{Ours}}
         & \multicolumn{3}{c}{\textbf{NeRF-A~\cite{martin-brualla_nerf_w_2021}}} \\
         \cmidrule(rl){1-3} \cmidrule(rl){4-6}
         \textbf{PSNR}$\uparrow$ & \textbf{SSIM}$\uparrow$ & \textbf{LPIPS}$\downarrow$ 
         & \textbf{PSNR}$\uparrow$ & \textbf{SSIM}$\uparrow$ & \textbf{LPIPS}$\downarrow$ \\
        \midrule
         33.08 & 0.951 & 0.154
         & 23.52 & 0.907 & 0.247
         \\
         \bottomrule
        \end{tabular}
    }
    \label{sup_table:control_render}
\end{wraptable}

In our experiments, we use NeRF-A (appearance), which means without transient part.
For controllable rendering, NeRF-A interpolates their appearance between each view, which has ambiguity in rendering results and cannot control explicitly.
In contrast, HDR-Plenoxels uses a tone-mapping module based on explicit functions and can control LDR rendering with quantified value input.
Our synthetic dataset contains three different exposures, and we conduct controllable rendering, which reconstructs median exposure given brighter and darker values or embeddings.
In our HDR-Plenoxels, we get minimum and maximum value of exposure after training and get median value for rendering median exposure.
In NeRF-A, we interpolate between appearances embeddings which are assigned to brighter and darker images, respectively.
We measure quantitative quality between them with median exposure ground truth images.
As shown in \Tref{sup_table:control_render}, ours can control radiometric calibration more accurately and also get precise geometry.\label{sec:C.3}

\section{Additional Experiments}\label{sec:D}
\subsection{Denoising Effects}
\begin{wrapfigure}{r}{0.5\linewidth}
  \setlength{\abovecaptionskip}{-1mm}
  \centering
  \includegraphics[width=1\linewidth]{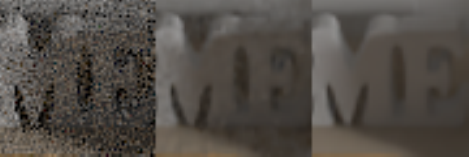}
  \caption{\textbf{Denoising of HDR-Plenoxels.}}
    \label{sup_fig:qual_denoise}
\end{wrapfigure}
We build our synthetic datasets using Blender\cite{blender}, which can render with or without the shot noise.
To verify the robustness of our model's novel view synthesis performance under such noise, we compare the PSNR results in the presence of the shot noise.
Our model marks 29.53 and $\ourpsnr$ in PSNR for kitchen data with and without shot noise, respectively.
Although shot noise degrades the numerical performance slightly (middle), it represents similar qualitative results to the model trained on denoised images (right), as shown in \Fref{sup_fig:qual_denoise}.
The left one is from the ground truth LDR image with shot noise.
The middle one is the novel view synthesis result.
As our model aggregates multi-view information, it somewhat shows a denoising effect.

\label{sec:D.1}
\subsection{Extreme Camera Conditions}
\noindent
\begin{wrapfigure}{r}{0.38\linewidth}
  \setlength{\abovecaptionskip}{-1mm}
  \centering
  \includegraphics[width=0.99\linewidth]{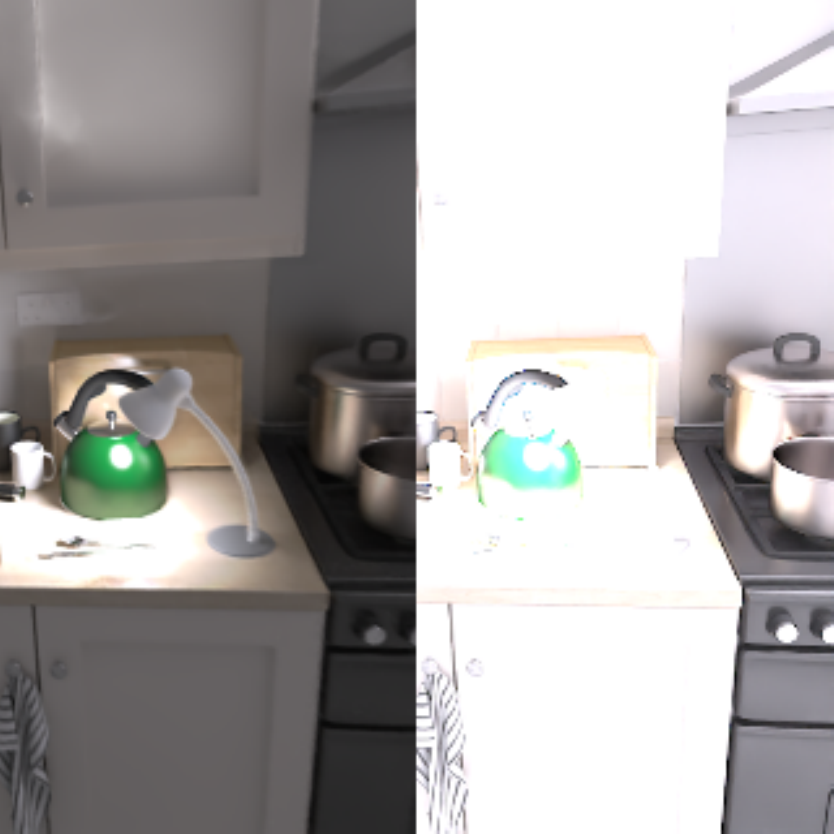}
  \caption{\textbf{Sample images of extreme exposure.}}
  \label{sup_fig:qual_extreme}
\end{wrapfigure}

Our original exposure setting has three levels in the $\pm$3EV range.
However, empirically, our model can robustly learn even under harsher exposure conditions.
For more extreme cases, \eg, very dark or bright, we train and evaluate our kitchen data with five exposure levels in respective $\pm$4EV, $\pm$5EV, and $\pm$6EV ranges.
For $\pm$3EV, $\pm$4EV, $\pm$5EV, and $\pm$6EV cases, ours obtains 31.58, 30.30, 29.10, and 28.48 in PSNR, respectively. 
Although PSNR steadily decreases as the exposure gap becomes wider, ours at the most extreme setting obtains higher PSNR than ADOP in the original setting (20.13).
Even in extreme conditions, our model shows high-quality HDR novel view synthesis (left) result with +6EV input LDR image (right) as shown in \Fref{sup_fig:qual_denoise}.
Our method is robust in various exposure settings, even in extreme cases.
\label{sec:D.2}
\subsection{Generality of the Tone Mapping Module}
\noindent
\begin{wrapfigure}{r}{0.675\linewidth}
  \setlength{\abovecaptionskip}{-1mm}
  \centering
  \includegraphics[width=0.99\linewidth]{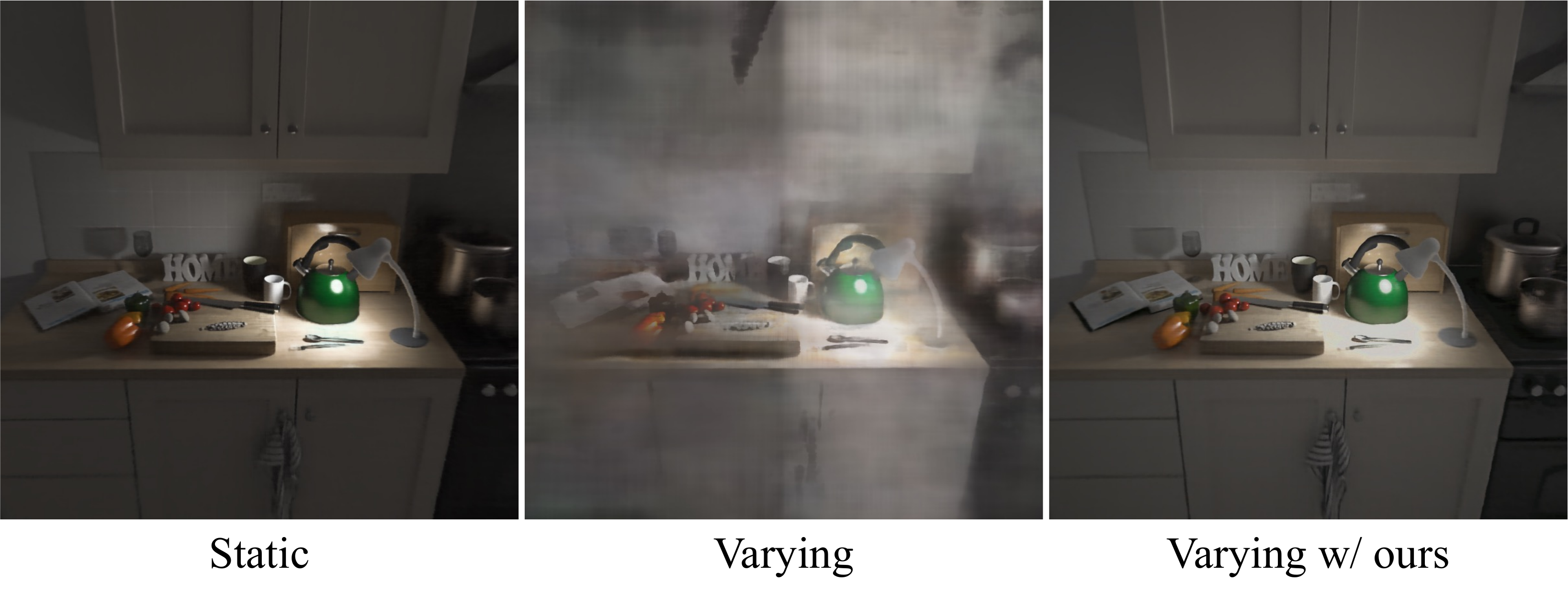}
  \caption{\textbf{NeRF with our tone mapping module.}}
  \label{sup_fig:qual_extreme}
\end{wrapfigure}

To verify the generality of the tone mapping module, we apply our tone mapping module to vanilla NeRF~\cite{nerf}.
We trained vanilla NeRF on our kitchen dataset.
Vanilla NeRF trained on images from varying cameras results in blurry and foggy images (middle).
NeRF with our tone-mapping module (right) shows clear novel view rendering results similar to a model trained with static camera settings (left).
Our tone mapping function enabled vanilla NeRF to learn radiance fields from varying cameras robustly.
As the tone mapping module is computationally light and easily attachable after the ray-marching, it can generally be employed in the various volume rendering models. 
\clearpage\label{sec:D.3}

%
%
\end{document}